\documentclass{ieeeaccess}
\usepackage{amsmath,amssymb,amsfonts}
\usepackage{cite}
\usepackage{algorithmic}
\usepackage{graphicx}
\usepackage{textcomp}
\usepackage[colorlinks,breaklinks=true,bookmarks=false]{hyperref}
\usepackage[table,xcdraw]{xcolor}
\usepackage{pifont}% for tick and x
\usepackage{longtable}
\newcommand{\cmark}{\ding{51}}%
\newcommand{\xmark}{\ding{55}}%
\usepackage{tabularx}
\usepackage{adjustbox}
\usepackage{array}
\usepackage{booktabs}
\usepackage{xltabular}
\usepackage{caption}
\usepackage{color}
\usepackage{soul}
\usepackage{cite}
\usepackage{siunitx}
\usepackage{textcomp}
\usepackage{graphicx}
\usepackage{gensymb}%
\usepackage{subcaption}
\usepackage{multirow}
\usepackage{arydshln}
\usepackage{soul}
\usepackage{lipsum}

\definecolor{darkred}{RGB}{200,0,0}
\def\pixel{p}

\def\R{\mathbb{R}}

\def\T{\mathbf{T}}
\def\K{\mathbf{K}}
\newcommand{\fromto}[2]{{#1}\rightarrow {#2}}
\definecolor{limegreen}{rgb}{0.3, 0.85, 0.3}

\def\ourxmark{\color{darkred}\xmark\color{black}}
\usepackage{mathtools}

\def\BibTeX{{\rm B\kern-.05em{\sc i\kern-.025em b}\kern-.08em
    T\kern-.1667em\lower.7ex\hbox{E}\kern-.125emX}}
    
\usepackage{pgfplots}
\usepackage{pgfplotstable}
\pgfplotsset{compat=1.7}
\usepackage{tikz}

\NewSpotColorSpace{PANTONE}
\AddSpotColor{PANTONE} {PANTONE3015C} {PANTONE\SpotSpace 3015\SpotSpace C} {1 0.3 0 0.2}
\SetPageColorSpace{PANTONE}

\usepackage{scalerel}
\usepackage{tikz}
\usetikzlibrary{svg.path}

\definecolor{orcidlogocol}{HTML}{A6CE39}
\tikzset{
  orcidlogo/.pic={
    \fill[orcidlogocol] svg{M256,128c0,70.7-57.3,128-128,128C57.3,256,0,198.7,0,128C0,57.3,57.3,0,128,0C198.7,0,256,57.3,256,128z};
    \fill[white] svg{M86.3,186.2H70.9V79.1h15.4v48.4V186.2z}
                 svg{M108.9,79.1h41.6c39.6,0,57,28.3,57,53.6c0,27.5-21.5,53.6-56.8,53.6h-41.8V79.1z M124.3,172.4h24.5c34.9,0,42.9-26.5,42.9-39.7c0-21.5-13.7-39.7-43.7-39.7h-23.7V172.4z}
                 svg{M88.7,56.8c0,5.5-4.5,10.1-10.1,10.1c-5.6,0-10.1-4.6-10.1-10.1c0-5.6,4.5-10.1,10.1-10.1C84.2,46.7,88.7,51.3,88.7,56.8z};
  }
}

\newcommand\orcidicon[1]{\href{https://orcid.org/#1}{\mbox{\scalerel*{
\begin{tikzpicture}[yscale=-1,transform shape]
\pic{orcidlogo};
\end{tikzpicture}
}{|}}}}

\makeatletter
\def\@BTrule[#1]{%
  \ifx\longtable\undefined
    \let\@BTswitch\@BTnormal
  \else\ifx\hline\LT@hline
    \nobreak
    \let\@BTswitch\@BLTrule
  \else
     \let\@BTswitch\@BTnormal
  \fi\fi
  \global\@thisrulewidth=#1\relax
  \ifnum\@thisruleclass=\tw@\vskip\@aboverulesep\else
  \ifnum\@lastruleclass=\z@\vskip\@aboverulesep\else
  \ifnum\@lastruleclass=\@ne\vskip\doublerulesep\fi\fi\fi
  \@BTswitch}
\makeatother

\definecolor{red0}{RGB}{100, 20, 20}
\definecolor{red1}{RGB}{180, 20, 20}
\definecolor{red2}{RGB}{250, 20, 20}
\definecolor{orange0}{RGB}{150, 100, 0}
\definecolor{orange1}{RGB}{200, 100, 0}
\definecolor{orange2}{RGB}{250, 100, 0}
\definecolor{purple0}{RGB}{150, 45, 160}
\definecolor{purple1}{RGB}{210, 45, 160}
\definecolor{purple2}{RGB}{250, 45, 160}
\definecolor{cyan0}{RGB}{0, 200, 160}
\definecolor{cyan1}{RGB}{0, 200, 200}
\definecolor{cyan2}{RGB}{0, 200, 240}
\definecolor{green0}{RGB}{0, 100, 0}
\definecolor{green1}{RGB}{0, 140, 0}
\definecolor{green2}{RGB}{0, 180, 0}
\definecolor{lightgreen0}{RGB}{200, 100, 100}
\definecolor{lightgreen1}{RGB}{200, 140, 100}
\definecolor{lightgreen2}{RGB}{200, 180, 100}

\begin{document}
\history{Date of publication xxxx 00, 0000, date of current version xxxx 00, 0000.}
\doi{10.1109/ACCESS.2017.DOI}
% HUMBI: A Large Multiview Dataset of Human Body Expressions
\title{HUMAN4D: A Human-Centric Multimodal Dataset for Motions \& Immersive Media}
\author{
\uppercase{Anargyros Chatzitofis}\authorrefmark{1,2} \orcidicon{0000-0002-3848-4210} \IEEEmembership{Graduate Student, IEEE},
\uppercase{
Leonidas Saroglou\authorrefmark{2} \orcidicon{0000-0003-1471-4779},
Prodromos Boutis\authorrefmark{2} \orcidicon{0000-0002-7518-1657},
Petros Drakoulis\authorrefmark{2} \orcidicon{0000-0003-3434-3290},
Nikolaos Zioulis\authorrefmark{2} \orcidicon{0000-0002-7898-9344},\\
Shishir Subramanyam\authorrefmark{3} \orcidicon{0000-0002-3653-7630},
Bart Kevelham\authorrefmark{4} \orcidicon{0000-0001-7814-5128},
Caecilia Charbonnier\authorrefmark{4} \orcidicon{0000-0002-7018-885X},
Pablo Cesar}\authorrefmark{3} \orcidicon{0000-0003-1752-6837} \IEEEmembership{Senior, IEEE},
\uppercase{Dimitrios Zarpalas\authorrefmark{2} \orcidicon{0000-0002-9649-9306},
Stefanos Kollias}\authorrefmark{1} \orcidicon{0000-0003-2899-0598} \IEEEmembership{Fellow, IEEE},
\uppercase{Petros Daras}\authorrefmark{2} \orcidicon{0000-0003-3814-6710} \IEEEmembership{Senior, IEEE}}

\address[1]{National Technical University of Athens, School of Electrical \& Computer Eng., Athens, Greece (tofis3d@central.ntua.gr (A.C), stefanos@cs.ntua.gr (S.K))}
\address[2]{Centre for Research \& Technology Hellas, Information Technologies Institute, Thessaloniki, Greece (tofis@iti.gr (A.C), saroglou@iti.gr (L.S), prod@iti.gr (P.B), petros.drakoulis@iti.gr (P.D), nzioulis@iti.gr (N.Z), zarpalas@iti.gr (D.Z), daras@iti.gr (P.D))}
\address[3]{Centrum Wiskunde \& Informatica. Amsterdam, Netherlands (s.subramanyam@cwi.nl (S.S), p.s.cesar@cwi.nl (P.S))}
\address[4]{Artanim Foundation, Geneva, Switzerland (bart.kevelham@artanim.ch (B.K), caecilia.charbonnier@artanim.ch (C.C))}

\tfootnote{This work was supported by the EU funded project VRTogether H2020 under the grant agreement 762111.\\
\textbf{Author Contributions - } 
A.C: Conceptualization, Methodology, Software, Validation, Formal Analysis, Investigation, Data Curation, Supervision; 
L.S: Methodology, Software, Validation, Formal Analysis, Investigation;
P.B: Methodology, Software, Formal Analysis, Investigation, Data Curation; 
P.D: Software, Validation, Formal Analysis, Investigation;
N.Z: Methodology, Software;
S.S: Methodology, Investigation, Data Curation;
B.K: Methodology, Data Curation, Resources;
C.C: Methodology, Data Curation, Resources;
P.C: Methodology;
D.Z: Supervision;
S.K: Supervision;
P.D: Supervision, Funding Acquisition;
}

\markboth
{Chatzitofis \headeretal: Preparation of Papers for IEEE TRANSACTIONS and JOURNALS}
{Chatzitofis \headeretal: Preparation of Papers for IEEE TRANSACTIONS and JOURNALS}

\corresp{Corresponding author: Anargyros Chatzitofis (e-mail: tofis3d@central.ntua.gr, tofis@iti.gr).}

\begin{abstract}
We introduce HUMAN4D, a large and multimodal 4D dataset that contains a variety of human activities simultaneously captured by a professional marker-based MoCap, a volumetric capture and an audio recording system.
By capturing $2$ female and $2$ male professional actors performing various full-body movements and expressions, HUMAN4D provides a diverse set of motions and poses encountered as part of single- and multi-person daily, physical and social activities (jumping, dancing, etc.), along with multi-RGBD (mRGBD), volumetric and audio data.
Despite the existence of multi-view color datasets captured with the use of hardware (HW) synchronization, to the best of our knowledge, HUMAN4D is the first and only public resource that provides volumetric depth maps with high synchronization precision due to the use of intra- and inter-sensor HW-SYNC.
Moreover, a spatio-temporally aligned scanned and rigged 3D character complements HUMAN4D to enable joint research on time-varying and high-quality dynamic meshes.
We provide evaluation baselines by benchmarking HUMAN4D with state-of-the-art human pose estimation and 3D compression methods.
We apply OpenPose and AlphaPose reaching 70.02\% and 82.95\% mAP\textsubscript{PCKh-0.5} on single- and 68.48\% and 73.94\% mAP\textsubscript{PCKh-0.5} on two-person 2D pose estimation, respectively.
In 3D pose, a recent multi-view approach named Learnable Triangulation, achieves 80.26\% mAP\textsubscript{PCK3D-10cm}.
For 3D compression, we benchmark Draco, Corto and CWIPC open-source 3D codecs, respecting online encoding and steady bit-rates between 7-155 and 2-90 Mbps for mesh- and point-based volumetric video, respectively.
Qualitative and quantitative visual comparison between mesh-based volumetric data reconstructed in different qualities and captured RGB, showcases the available options with respect to 4D representations.
HUMAN4D is introduced to enable joint research on spatio-temporally aligned pose, volumetric, mRGBD and audio data cues.
The dataset and its code are available \href{https://tofis.github.io/myurls/human4d}{online}.
\end{abstract}


\titlepgskip=-15pt

\maketitle

\section{Introduction}\label{sec::introduction}

Inhabitance in a 4D world of moving 3D objects of various shapes and colors increases the need to capture and extensively study, analyze and exploit the 4D data around us, especially now, with the massive development of low-cost sensing devices \cite{mukherjee2020beginner}.
Nowadays, volumetric video of humans, captured with the aid of multiple cameras, and scanned 3D characters, animated with the use of motion capture (MoCap) technologies, comprise the core elements for human-centric 4D media production, a domain essential in several technological and industrial sectors.

On the one hand, these technologies constitute key elements in immersive experiences that provide remote virtual presence and co-presence (e.g. XR conferencing \cite{gunkel2018virtual}, XR museums \cite{lee2020experiencing}, etc.). 
The experiences are further enhanced by augmenting the virtual and immersive worlds with photorealistic representations that enable highly natural and realistic audiovisual communication between multiple users.

On the other hand, dense 4D data cues produced with such technologies contain space-time coherent information of shape, motion, and appearance of people, attracting the interest of the computer vision research community and beyond.
Several research works \cite{bogo2017dynamic, AMASS:2019} provide large corpora with synthetic humans generated based on human body priors \cite{loper2015smpl}, motion capture data and more.
By applying 3D surface reconstruction methods \cite{newcombe2015dynamicfusion, alexiadis2016integrated, dou2016fusion4d, orts2016holoportation, jackson20183d, guo2017real, alldieck2018video, guo2019relightables} on 3D or 4D data captured with single or multiple spatio-temporally aligned RGBD sensors, volumetric video is reconstructed in either real-time or offline.
Fusing volumetric video with high quality 3D scans and motion capture enables the study and development of data-driven approaches across several domains, such as 2D human pose estimation \cite{cao2017realtime, he2017mask, alp2018densepose, guler2019holopose, kolotouros2019learning}, 3D pose estimation \cite{mehta2017vnect, cao2018openpose, carraro2018real, iskakov2019learnable, qiu2019cross, chatzitofis2019deepMoCap, tripathi2020posenet3d}, motion analysis \cite{alexiadis2014quaternionic, patrona2018motion}, 3D/4D volumetric reconstruction \cite{kordelas2010state, newcombe2015dynamicfusion, alexiadis2016integrated, dou2016fusion4d, orts2016holoportation, jackson20183d, guo2017real, alldieck2018video}, performance capture \cite{habermann2019livecap, habermann2020deepcap}, volumetric video compression \cite{doumanoglou2014toward, mekuria2016design, quach2019learning, wang2019learned, tang2020deep}, photorealistic representations \cite{guo2019relightables} and more.

The advancement of shape and motion computer vision techniques, the development of immersive media technologies, as well as the interest of the industry in human-centric 4D media production, highly and rapidly increase the need for large, high-quality datasets that will act as cornerstones for their continuous development, also enabling their joint evolution.
Nevertheless, at the moment, only few datasets are partially focused on some of the aspects of these challenging tasks.

On top of that, several computer vision methods approach 3D/4D research tasks from monocular or HW-SYNCed multi-view color (i.e. 2D) streams.
However, by definition, 2D data cannot cope with the intricacies of 3D/4D shape or form, at least to the extent that the volumetric data can.
That is probably due to the lack of HW-SYNCed depth/volumetric data from public resources. 
For instance, the lack of HW-SYNCed volumetric data along with ground-truth 3D poses for supervision eliminates the attempts for data-driven 3D pose estimation approaches from volumetric data.

To this end, we create HUMAN4D, a dataset that fills these gaps by providing professional motion capture along with volumetric data captured in 3D character and mesh- and point-based volumetric representations.
In particular:
\begin{itemize}
    \item We introduce a publicly available 4D dataset containing a large corpus of annotated spatio-temporally aligned multi-view RGBD (mRGBD), volumetric and motion capture data, in order to enable extensive research on several computer vision and graphics topics.
    \item To the best of our knowledge, HUMAN4D is the first dataset that provides HW-SYNCed mRGBD frames along with marker-based motion capture and audio data cues, with the use of recent consumer-grade depth sensing devices, cutting-edge optical motion capture technologies and body-worn audio recording, respectively.
    \item We provide pose estimation baselines by applying data-driven 2D and 3D pose estimation algorithms on single- and multi-view data sequences, along with insights with respect to the advantages of HUMAN4D for training such methods.
    \item We perform and report a detailed study on volumetric data compression using 3D codecs, examining the rate distortion from several perspectives, while respecting online volumetric video encoding and steady bit-rates.
    \item We conduct and report objective visual quality evaluation on various volumetric representations, i.e. mesh-based volumetric data evaluation across various reconstruction qualities.
\end{itemize}
The remainder of this paper is organized as follows: 
Sec. \ref{sec::related_work} overviews related datasets including 4D data in a similar aspect; 
Sec. \ref{sec::dataset} describes in detail the HUMAN4D dataset, giving evidence with respect to its creation and statistics; 
Sec. \ref{sec::bench::MoCap} benchmarks 2D and 3D pose estimation data-driven models on HUMAN4D; while Sec. \ref{sec::bench::volvideo} benchmarks 3D codecs and compares mesh-based 4D representations with respect to visual quality using well-known objective metrics;
in Sec. \ref{sec::discussion}, we discuss the impact of this dataset to the research community and beyond;
finally, Sec. \ref{sec::conclusion} concludes the paper and discusses future work.
 
%%%%%%%%%%%%%%%%%%%%%%%%%%%%%%%%%%%%%%%%%%
\section{Related work}\label{sec::related_work}

\begin{table*}[]
\centering
\begin{adjustbox}{width=0.85\textwidth}
\begin{tabular}{ lccccccc }
\toprule
 & \textbf{MHAD}\textsubscript{(2013)} \cite{ofli2013berkeley} &  \textbf{Human3.6M}\textsubscript{(2014)} \cite{h36m_pami} &
\textbf{CMUPanoptic}\textsubscript{(2015)} \cite{joo2015panoptic} & \textbf{HUMBI}\textsubscript{(2018)} \cite{yu2018humbi} 
& \textbf{HUMAN4D}\textsubscript{(2020)} \\
\midrule
 \textit{Body Pose} 
 & \cmark 
 & \cmark  
 & \cmark 
 & \cmark 
 & \cmark \\
\hdashline
  \textit{Marker-based MoCap} 
    & \cmark 
  & \cmark  
  & \ourxmark 
  & \ourxmark  
  & \cmark \\
\hdashline
 \textit{Body Part Segments} 
  & \ourxmark  
 & \ourxmark 
 & \ourxmark 
 & \cmark 
 & \ourxmark  \\ 
\hdashline
  \textit{Multi-view RGB}
  & \cmark  
  & \cmark 
  & \cmark 
  & \cmark 
  & \cmark  \\ 
\hdashline
\textit{Multi-view Depth} 
& \cmark 
& \ourxmark  
& \cmark 
& \ourxmark 
& \cmark \\
\hdashline
 \textit{3D Meshes}  
 & \ourxmark 
 & \ourxmark 
 & \ourxmark 
 & \cmark 
 & \cmark \\
\hdashline
 \textit{Point-clouds} 
 & \ourxmark 
 & \ourxmark 
 & \cmark 
 & \ourxmark 
 & \cmark \\ 
\hdashline
 \textit{Audio Cues}  
 & \cmark 
 & \ourxmark 
 & \ourxmark 
 & \ourxmark 
 & \cmark \\ 
\hdashline
 \textit{Gaze Features}  
 & \ourxmark  
 & \ourxmark 
 & \ourxmark 
 & \cmark 
 & \ourxmark \\
\hdashline
 \textit{Hand Features} 
 & \ourxmark 
 & \ourxmark 
 & \cmark 
 & \cmark 
 & \ourxmark \\ 
\hdashline
 \textit{Facial Features} 
  & \ourxmark 
 & \ourxmark 
 & \cmark 
 & \cmark 
 & \ourxmark \\
\hdashline
 \textit{Rigged Characters} 
 & \ourxmark 
 & \cmark 
 & \ourxmark 
 & \ourxmark 
 & \cmark \\
\hdashline
 \textit{Multi-person}
  & \ourxmark 
 & \ourxmark 
 & \cmark 
 & \ourxmark 
 & \cmark \\ 
\bottomrule
\end{tabular}
\end{adjustbox}
\caption{Summary of state-of-the-art datasets and HUMAN4D with respect to the available features and modalities.}\label{tbl::dataset_modalities}
\end{table*}

Over the past few decades, the computer vision research community has showed an increased interest for virtual human related technologies. 
A variety of traditional and learning-based computer vision methods are targeting open research problems using motion, volumetric, image and action-based data. 
In this section, we discuss relevant datasets \cite{h36m_pami, sigal2010humaneva, joo2015panoptic, yu2018humbi, ofli2013berkeley, andriluka14cvpr}, providing details and explaining the nature of the data they offer to the research community.
A brief overview of these datasets follows, while Table \ref{tbl::dataset_modalities} summarizes their features and modalities.

\noindent \textit{\textbf{MHAD}}\cite{ofli2013berkeley}:
One of the first publicly available datasets offering MoCap and RGBD data is (Berkeley) MHAD.
The MHAD dataset contains spatio-temporally aligned data cues captured with a professional MoCap system with active markers \cite{phasespace_cite} along with $12$ RGB and $2$ MS Kinect v2 (RGBD) cameras, $6$ wearable inertial sensors (accelerometers only) and $4$ microphones, recording the audio signals during the performance of the actions. 
The dataset consists of $659$ data sequences from $11$ human actions performed by $12$ subjects.
Although MHAD enables research on multi-view pose estimation and beyond, the MS Kinect v2 devices are only 2 and not HW-SYNCed, resulting in the existence of spatio-temporal offsets between the deprojected depth maps (point-clouds) and the 3D poses of the MoCap, limiting that way the joint use of 3D pose and volumetric data.

\noindent \textit{\textbf{Human3.6M}} \cite{h36m_pami}:
Human3.6M (H36M) contains a huge corpus with $3.6$ million 3D human poses of 5 female and 6 male subjects.
Similarly to HUMAN4D, the subjects perform a set of motions and poses (captured with $10$ motion capture cameras) from daily human activities (taking photos, talking on the phone, eating, sitting, etc.), along with synchronized color images from $4$ synchronized color cameras, depth maps from $1$ single Time-of-Flight (ToF) depth sensor and accurate 3D body scans of the subject actors involved.
H36M constitutes one of the most widely used datasets for human-centric computer vision research tasks, however, there still exist some drawbacks.
Only the color cameras support hardware inter-synchronization, there is only one depth sensor with low depth map resolution, while the set of motion capture cameras is limited (10)  in comparison with HUMAN4D (24).
Finally, the recent human-centric research advances and efforts are focused on multi-person captures (e.g. including social activities) similar to ones provided by HUMAN4D and other datasets \cite{joo2015panoptic, yu2018humbi}, contrary to H36M which contains only single-person sequences.

\noindent \textit{\textbf{CMUPanoptic}}\cite{joo2015panoptic}:
CMUPanoptic (CMU) is the largest public dataset in terms of the number of camera views (521), capturing natural interactions of up to 8 subjects performing social activities with uncontrolled behaviour and appearance.
The dataset has been captured using the Panoptic Studio \cite{joo2015panoptic}, a massively multi-view capture system consisting of $480$
VGA, $31$ HD and $10$ RGBD (Kinect v2) cameras, distributed over the surface of a geodesic sphere.
Beyond body poses, CMU also contains 3D facial landmarks and 2D/3D hand pose data cues.
Even though CMU currently constitutes one of the richest publicly available datasets in the field, HUMAN4D enables further research perspectives.
Despite its spatio-temporal setting, CMU does not provide HW volumetric synchronization since the time alignment between the Kinect v2 RGBD streams is achieved through a hardware modification using the microphone array of each device, incapable to provide synchronization precision comparable to HUMAN4D (see Sec. \ref{sec::st-mrgbd}).
Finally, the pose estimates have not been captured using a professional marker-based motion capture solution as in HUMAN4D; instead, an accurate marker-less approach has been used.

\noindent \textit{\textbf{HUMBI}}\cite{yu2018humbi}:
Another large and publicly available multi-view dataset is HUMBI, focusing on human body expressions with natural clothing, aiming to facilitate modeling of view-specific appearance and geometry of gaze, face, hand, body, and garment from several and various people.
HUMBI complements the publicly available datasets with respect to the number of camera views ($107$ synchronized HD cameras) and subjects ($772$ distinctive subjects across gender, ethnicity, age, and physical condition).
The dataset includes five elementary body expressions, i.e. gaze, face, hand, body and garment.
With the use of SMPL\cite{loper2015smpl}, HUMBI provides mesh-based 3D geometry of the subjects along with their respective texture atlases.
For HUMBI, the use of depth sensors was out of scope, thus multi-view depth sensing was not considered.

HUMAN4D aims to tackle lacking areas of existing, publicly available 4D datasets. 
HUMAN4D consists of a large corpus of spatio-temporally aligned mRGBD, volumetric and motion capture data cues, providing high synchronization precision between the multiple RGBD streams exploiting the HW-SYNC capabilities of the sensors.
On top of that, HUMAN4D contains (social) activities between multiple subjects (2), enabling research on challenging computer vision tasks under the multi-person aspect (e.g. occlusions, multiple person instances in the field of view, larger volumetric areas, etc.).
HUMAN4D is meant to provide the computer vision research community with data that will enable the research and development of novel approaches on intensively active human-centric research domains.
It is worth noting that the consumer-grade depth sensing devices used for the RGBD data capturing are commercially available in the market, allowing the experimentation and development of computer vision algorithms applicable even for production purposes.

%%%%%%%%%%%%%%%%%%%%%%%%%%%%%%%%%%%%%%%%%%
\section{HUMAN4D Dataset}\label{sec::dataset}

\begin{figure*}[t]
\begin{center}\includegraphics[width=\textwidth]{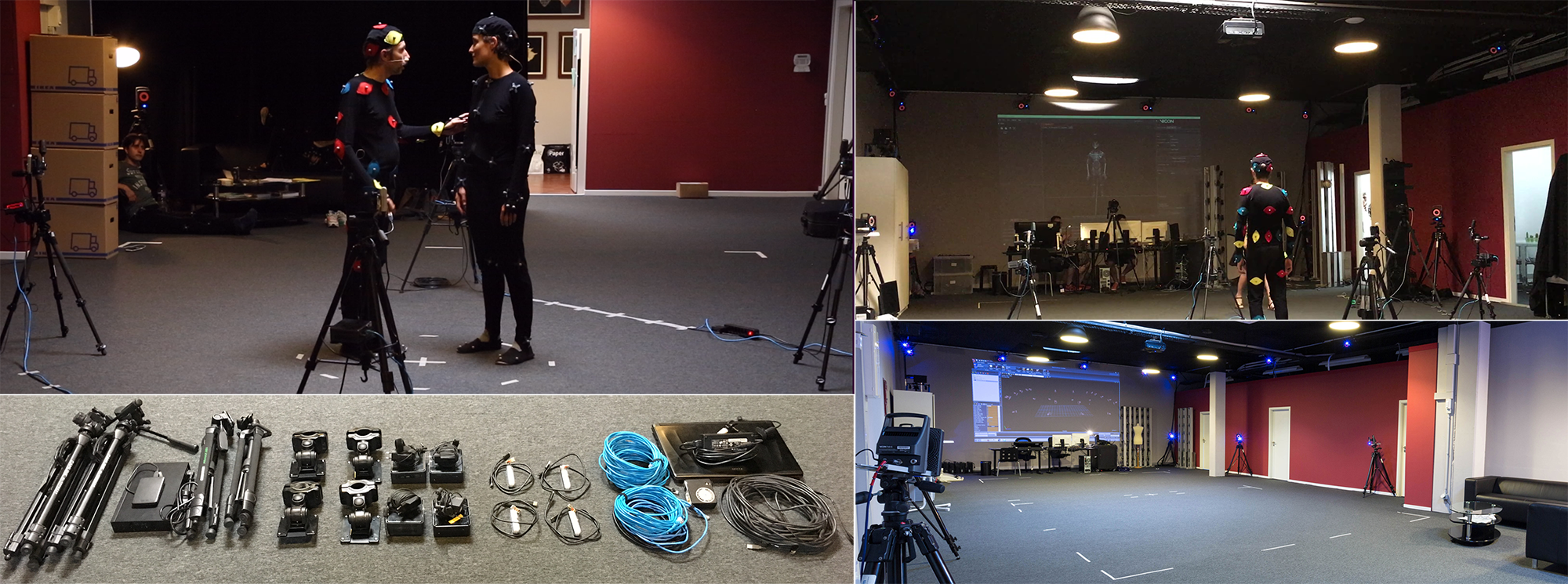}
\caption{Pictures taken during the preparation and capturing of the HUMAN4D dataset (in Artanim's facilities). 
The room is equipped with 24 Vicon MXT40S cameras  rigidly placed on the walls, a portable volumetric capturing system with 4 Intel RealSense D415 depth sensors temporarily set up to capture the RGBD data cues and wearable microphones for the actors.}
\label{fig::capturing_space}
\end{center}
\end{figure*}

\subsection{4D Capturing Setting}\label{sec::dataset::setting}

\begin{figure}[h]
\begin{center}
\includegraphics[width=\columnwidth]{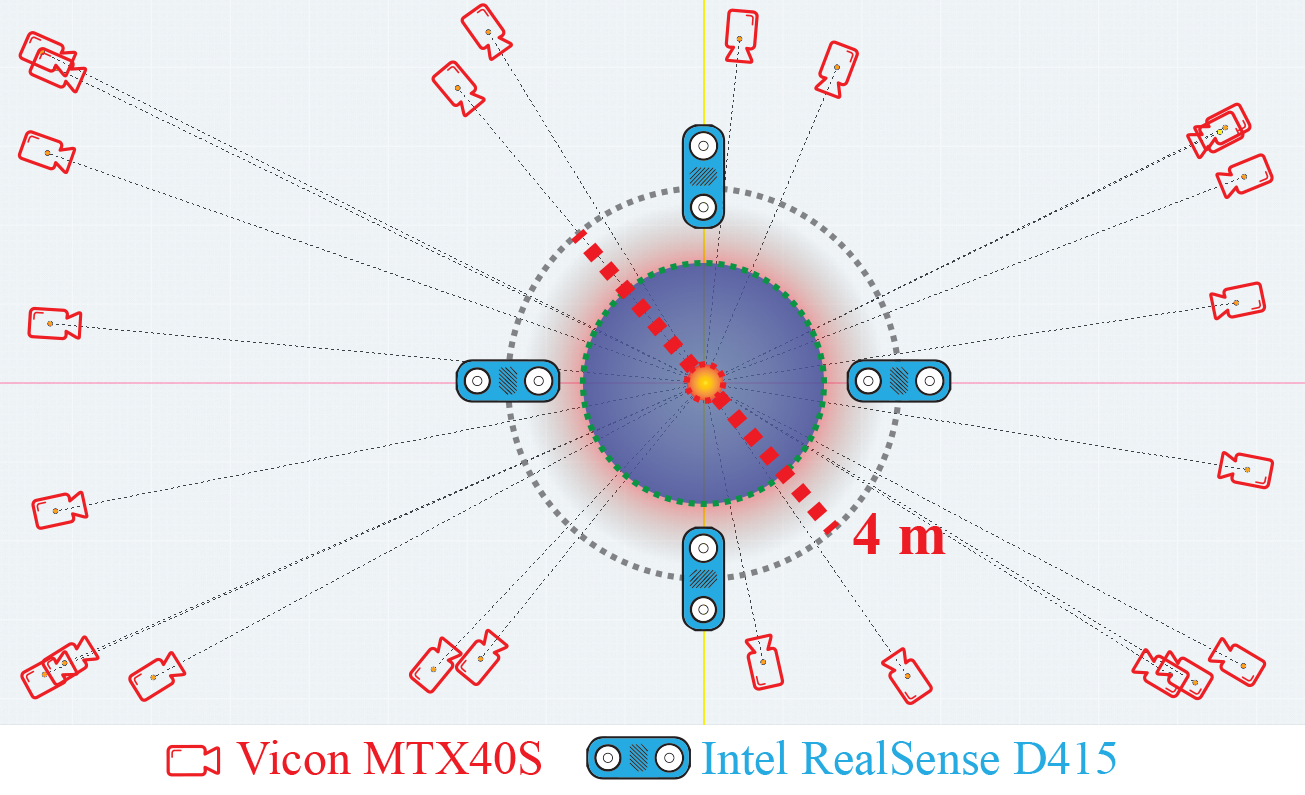}
\caption{Capturing space floor-plan showing the poses of $24$ Vicon MXT40S cameras and $4$ Intel RealSense D415 sensors.}
\label{fig::capturing_space_floor}
\end{center}
\end{figure}

The capturing of the dataset took place in a professional motion capture studio (Artanim Foundation\footnote{\url{http://artanim.ch/}}) where, beyond the motion capture system, special portable equipment for volumetric capturing was set up, as depicted in Fig. \ref{fig::capturing_space}.
In particular, 24 motion capture (MoCap) cameras along with 4 stereo-based depth sensors and microphones using HW and software (SW) synchronization (see Sec. \ref{sec::dataset::sync_and_calib} for details) were used, to capture the whole dataset.
All 24 motion capture cameras were rigged on the walls, to maximize the effective experimentation volume.
The high number of motion cameras (24) increases the accuracy of the motion capture due to the elimination of occlusions, providing that way high precision ground-truth poses for the dataset.   
The actual capturing space was set in an area of approximately $4m \times 4m$ so that the bodies of the actors were at least partially in the field-of-view of the RGBD cameras during the performances. 
These cameras were placed at the 4 corners of the stage in a cross schema. 
The floor-plan of the whole capturing setup is illustrated in Fig. \ref{fig::capturing_space_floor}.
Finally, a 3D body scanner was used to obtain an accurate 3D mesh-based volumetric model of one of the actors.

\begin{figure*}[t]
\begin{center}
\includegraphics[width=\textwidth]{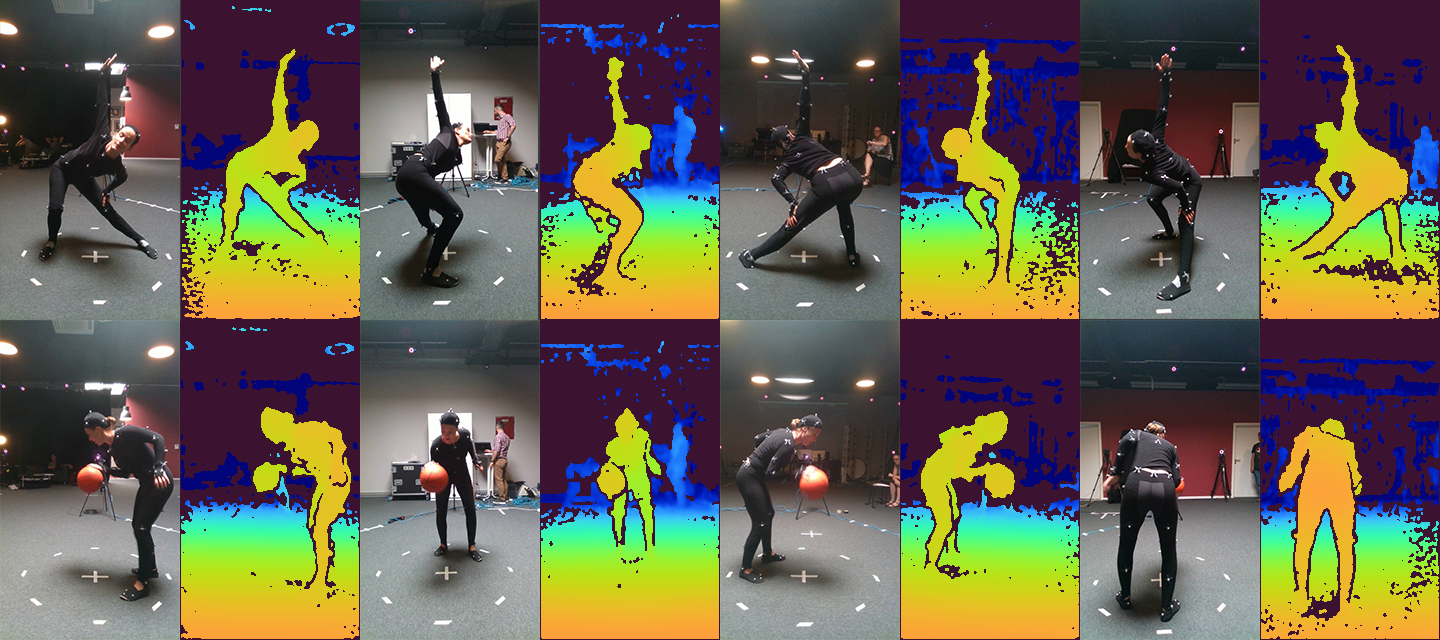}
\caption{HW-SYNCed multi-view RGBD samples (4 RGBD frames each) from \textit{"stretching\textunderscore n\textunderscore talking"} (top) and \textit{"basketball\textunderscore dribbling"} (bottom) activities. 
The depth maps are colorized using TURBO colormap \cite{Turbocolormap2019}.}
\label{fig::rgbd2}
\end{center}
\end{figure*}

\subsection{Dataset Creation}\label{sec::dataset::creation}

For the creation of the dataset, $4$ professional actors, $2$ female and $2$ male were recruited, in order to pursue the highest possible quality of the captured actions, with respect to the authenticity of the performances.
Within HUMAN4D, without the post-processing products (i.e. volumetric data), we captured and introduce the following:

\begin{table}[ht]
\centering
\caption{Details with respect to HUMAN4D physical, daily and social activities.}
\label{tbl::activities}
\begin{adjustbox}{width=\columnwidth}
\begin{tabular}{clccc}
\toprule
            & \textbf{activity} 
            & \textbf{\# frames} & \textbf{audio} & \textbf{type} \\
            \midrule
\parbox[t]{2mm}{\multirow{14}{*}{\rotatebox[origin=c]{90}{\textit{Single-person}}}}
& \textit{running}              
& 2,050                                     & \ourxmark                & physical                   \\
& \textit{jumping\textunderscore jack}                               
& 1,974                                     & \ourxmark                & physical                   \\
& \textit{bending}                                                           
& 2,156                                       & \ourxmark              & physical                      \\
& \textit{punching\textunderscore n\textunderscore kicking}                                            
& 2,079                                       & \ourxmark                & physical                    \\
& \textit{basketball\textunderscore dribbling}  
& 2,124                                       & \ourxmark              & physical                     \\
& \textit{laying\textunderscore down}  
& 4,082                                       & \ourxmark                       & physical            \\
& \textit{sitting\textunderscore down}   
& 3,288                                       & \ourxmark                  & daily         \\
& \textit{sitting\textunderscore on\textunderscore a\textunderscore chair}              
& 2,797                                       & \ourxmark           & daily                        \\
& \textit{talking}              
& 2,377                                       & \checkmark                                 & daily  \\
& \textit{object\textunderscore dropping\textunderscore n\textunderscore picking} 
&  1,768                                      & \ourxmark                        & daily           \\
& \textit{stretching\textunderscore n\textunderscore talking}                        
&  2,787                                      & \checkmark                    & physical               \\
& \textit{talking\textunderscore n\textunderscore walking} 
&  2,889                                      & \checkmark               & daily                    \\
& \textit{watching\textunderscore scary\textunderscore movie}          
& 2,194                                       & \checkmark               & daily                    \\
& \textit{in-flight\textunderscore safety\textunderscore announcement}  
&  6,192                                      & \checkmark              &       daily               \\
\midrule
\parbox[t]{2mm}{\multirow{5}{*}{\rotatebox[origin=c]{90}{\textit{Multi-person}}}}
& \textit{watching\textunderscore football\textunderscore together}  
&  1,760                                      & \checkmark          & social                         \\
& \textit{dancing\textunderscore together}                                    
& 1,356                                       & \checkmark            & social                         \\
& \textit{physical\textunderscore examination}                         
& 2,328                                       & \checkmark           & social                          \\
& \textit{whispering}   
&  3,045                                      & \checkmark            & social                         \\
& \textit{card\textunderscore trick} 
&  3,060                                      & \checkmark            & social        \\
\midrule
 &                                                    &  50,306                                      &            &   \\
\bottomrule
\end{tabular}
\end{adjustbox}
\end{table}

\begin{itemize}
    \item Multimodal data of $14$ single-person and $5$ two-person actions ($19$ in total), including physical exercises, daily and social activities, totalling $56$ single-person and $10$ two-person sequences, respectively.
    In Table \ref{tbl::activities}, details with respect to HUMAN4D activities are figured.
    \item Projection matrices and external calibration camera parameters retrieved using an anchor-based calibration method to reduce pairwise accumulating errors, enabling 2D projection of 4D data to the various camera views and vice versa.
    \item $30$ audio cues for some of the activities where the actors had to talk and act based on specific scripts and scenarios (see Table \ref{tbl::activities}).
    \item Synchronization between the modalities by providing timestamped data. 
    \item $1$ scanned and rigged 3D model of one of the professional actors.
    \item A set of benchmarks to facilitate comprehensive evaluation of 2D and 3D pose estimation methods, along with evaluation of volumetric video production and compression quality.
\end{itemize}

Following, we describe in detail the modalities we used and the techniques we applied to capture and create the dataset.

\subsubsection{SPATIO-TEMPORALLY ALIGNED mRGBD CAPTURE}\label{sec::st-mrgbd}
To the best of our knowledge, HUMAN4D is the first publicly available dataset that offers HW synchronized multi-view RGBD data captured in a real-time manner. 
Most of the existing datasets use synchronized RGB cameras \cite{h36m_pami} or previous versions of Microsoft Kinect for RGBD capturing \cite{joo2015panoptic}, which do not support HW triggering, requiring SW-based soft synchronization solutions.

\begin{figure*}[!htb]
\begin{center}
\includegraphics[width=\textwidth]{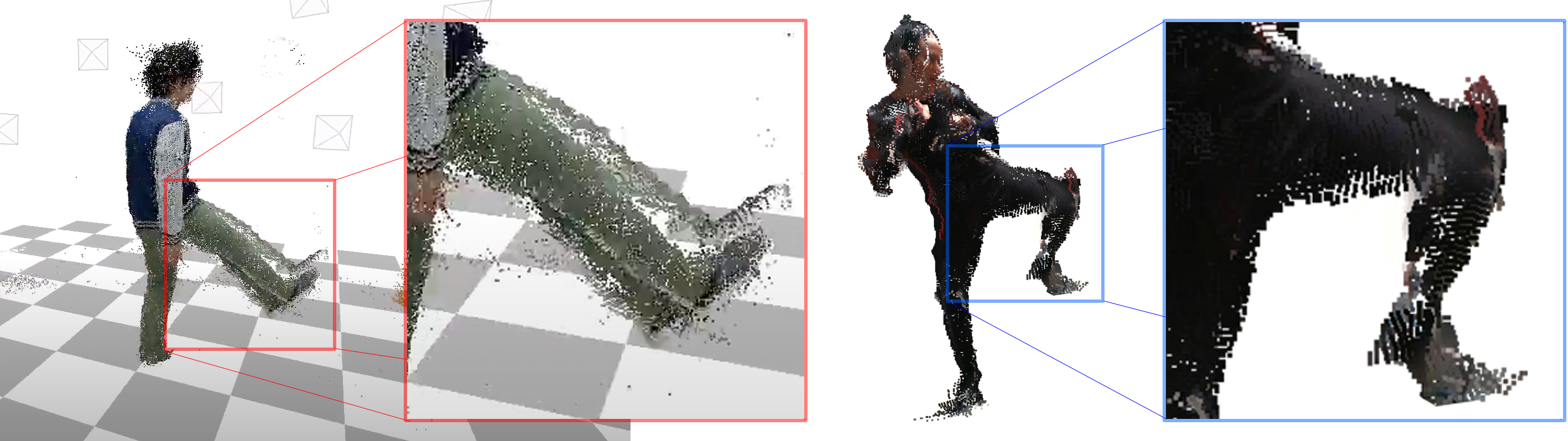}
\caption{Colored point-clouds from CMU \cite{joo2015panoptic} (Left) and HUMAN4D (Right) datasets showcase the benefits of HW-SYNC. In CMU, where the Kinect devices are modified for synchronization purposes, the leg of the subject is corrupted in a slow movement (i.e. slow leg lifting) due to the existence of temporal offsets between the devices. In HUMAN4D, the leg is appropriately captured in a fast movement (i.e. punching and kicking).}
\label{fig::syncing_issues}
\end{center}
\end{figure*}

In HUMAN4D, we instead use the Intel RealSense D415 sensor which offers this functionality \cite{grunnet2018using}.
D415 sensors can be configured in either master or slave synchronization mode, eliminating the need for external HW triggering when connected in a device cluster.
One device can be set as "master", providing the synchronization signal, and the rest as "slaves" that receive it and cohere.
The impact of HW-SYNCed mRGBD capture for volumetric- and pose-related tasks is depicted in Fig. \ref{fig::syncing_issues}, where point-clouds extracted by deprojecting mRGBD frames from HUMAN4D and CMU \cite{joo2015panoptic} are compared, showcasing the improved temporal alignment of the HW-SYNCed HUMAN4D against CMU data.
It is worth noting that CMU constitutes currently the only existing dataset that provides synchronized depth maps by applying a HW modification on the Kinect v2 devices.

Regarding depth capturing, the sensors were used in "high accuracy” mode, offering only the high confidence depth estimates, therefore producing accurate but sparse depth data.
It is worth noting that we configured the sensors exploiting their spatial filtering and exposure adjustment capabilities to capture the best possible depth quality.
We captured the mRGBD data using the capturing system\footnote{\url{https://github.com/VCL3D/VolumetricCapture}} proposed by Sterzentsenko \textit{et al.} \cite{sterzentsenko2018low}, while spatial alignment between  the  sensors  was  achieved  using  the  multi-sensor calibration schema proposed by Papachristou \textit{et al.} \cite{papachristou2018markerless}.
HW-SYNCed mRGBD samples are depicted in Fig. \ref{fig::rgbd2}.

\subsubsection{3D SCANNED AND RIGGED CHARACTER}

\begin{figure}[t]
\centering
 \includegraphics[width=\linewidth]{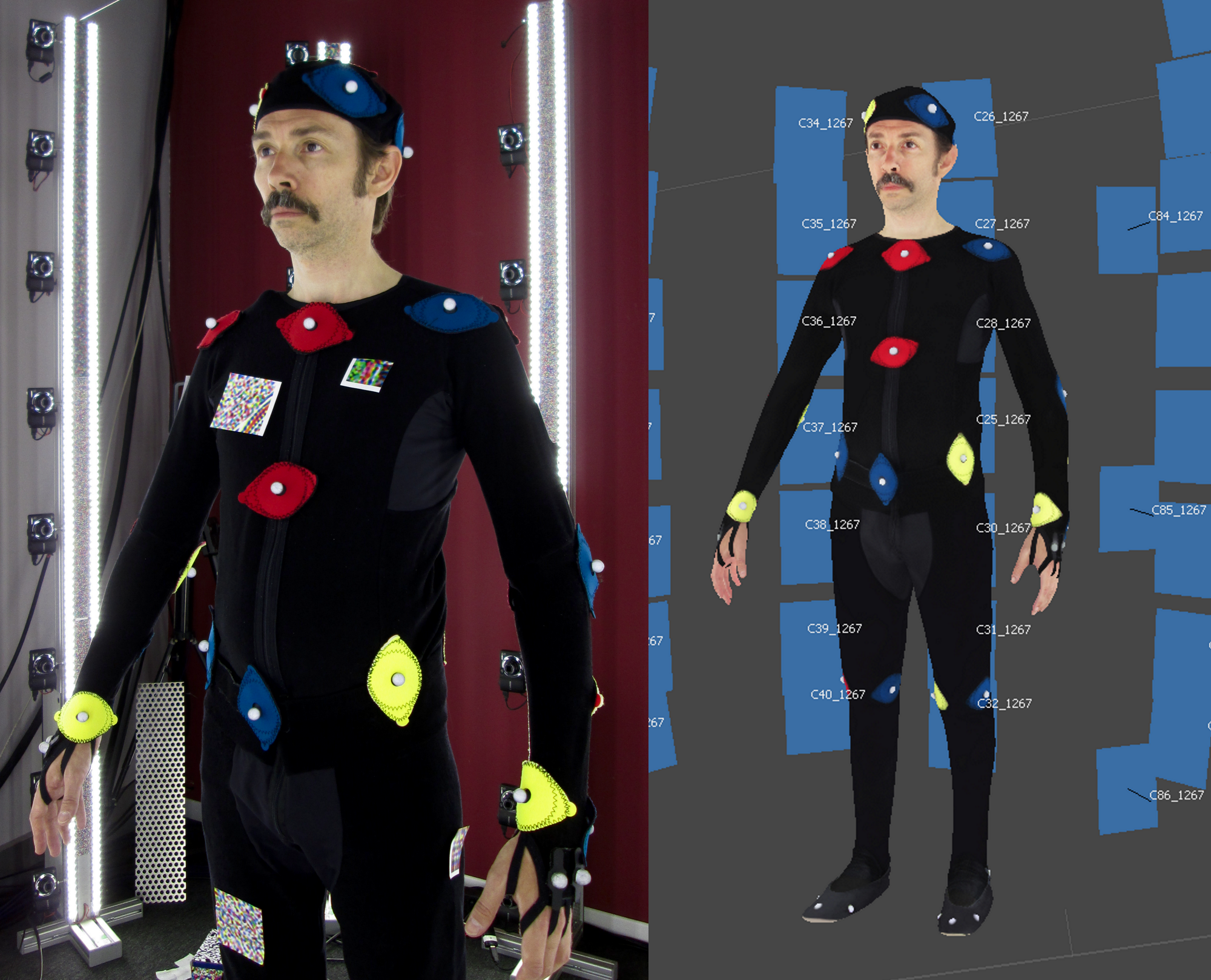}
%%%%%%%%%%%%%%%%%%%
\caption{Using a custom photogrammetry rig with 96 cameras, photos were taken of the actor (left) and reconstructed into a 3D textured mesh using Agisoft Metashape \cite{Agisoft152} (right).}
\label{fig::actor_bodyscan}
\end{figure}

To obtain an animatable mesh, one of the actors was scanned using a custom photogrammetry-based body scanning rig (Fig. \ref{fig::actor_bodyscan}). 
The rig consisted of 96 Canon Powershot A1400 cameras controlled using SW-based on the Canon Hack Development Kit (CHDK) \cite{CHDK2020}. 
Lighting was provided by LED strips mounted on the rig. 
All cameras were triggered in a synchronized manner. 
To aid the photogrammetric reconstruction of the bodyscan, the dark MoCap suit worn by the actor was temporarily augmented with colored paper markers, which were removed before the MoCap process. 

Using a commercial photogrammetry SW tool, Agisoft Metashape \cite{Agisoft152}, the individual photos were aligned to reconstruct a textured 3D mesh.
After the cleanup of mesh artifacts from the reconstruction process, the mesh was rigged and skinned for animation, using a standard full-body humanoid skeleton created by a professional 3D animator.  

\subsubsection{OPTICAL MARKER-BASED MOTION CAPTURE}\label{sec::dataset::MoCap}
To obtain reference animation of the 4 actors performing the various activities, a professional motion capture setup was used. 
The setup consisted of 24 Vicon MXT40S cameras (Vicon, Oxford Metrics, UK) sampling at 120Hz. 
Each actor wore a dedicated motion capture suit with $53$ attached retro-reflective markers.
This dense marker set along with the high number of motion cameras (24) allowed us to capture highly accurate and precise MoCap data to serve as ground-truth for training, supervising and evaluating data-driven approaches and beyond.  

For the purpose of subject calibration, each actor was asked to perform a full range of motion of all joints. 
The procedure ensured that the joint locations were correctly mapped to the set of the tracked markers.
Before each activity, the actors were asked to start in a T-pose and then proceed to their assigned activity.

\begin{figure}[!htb]
\centering
 \includegraphics[width=\linewidth]{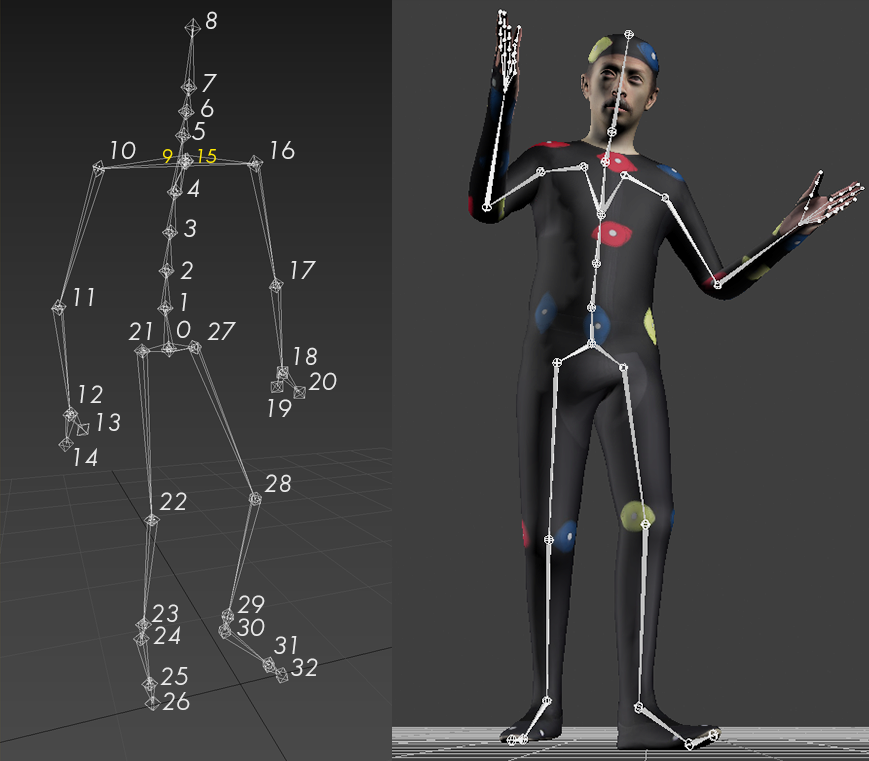}
%%%%%%%%%%%%%%%%%%%
\caption{(Left) Initial MoCap skeleton structure mapped to 3D and 2D pose joint indices. (Right) Animations for the scanned actor were re-targeted to match the skeleton rig of the 3D model.}
\label{fig::actor_rig_retarget}
\end{figure}

The captured animations of the actor whose body was subsequently scanned, underwent a retargeting process by a professional 3D animator. 
The goal of this process was to adjust the recorded animations to where slight differences between the captured MoCap skeleton structure and the one of the rigged 3D model exist, as illustrated in Fig. \ref{fig::actor_rig_retarget} (Right). 

\subsubsection{AUDIO RECORDING}\label{sec::audio}
The use of audio and its fusion with visual data have shown significant results in various research tasks such as human emotion recognition \cite{hossain2019emotion}, scene analysis \cite{owens2018audio}, human activity recognition \cite{subedar2019uncertainty} and more. 
To this end, also targeting the capture of social activities, we recorded audio during the performance of some of the actions. 
In particular, $30$ of the activities (see Table \ref{tbl::activities}) include audio either as a monologue (single-person) or conversation between two subjects, based on the related scripts and scenarios.
For this purpose, wireless body-worn microphones were used to record the audio cues.
The audio recording was performed at the frequency of $48$ kHz.

%%%%%%%%%%%%%%%%%%%%%%%%%%%%%%%%%%%%%%%%%%
\subsection{Dataset processing and annotations}

\begin{figure*}[!htb]
\begin{center}
\includegraphics[width=\textwidth]{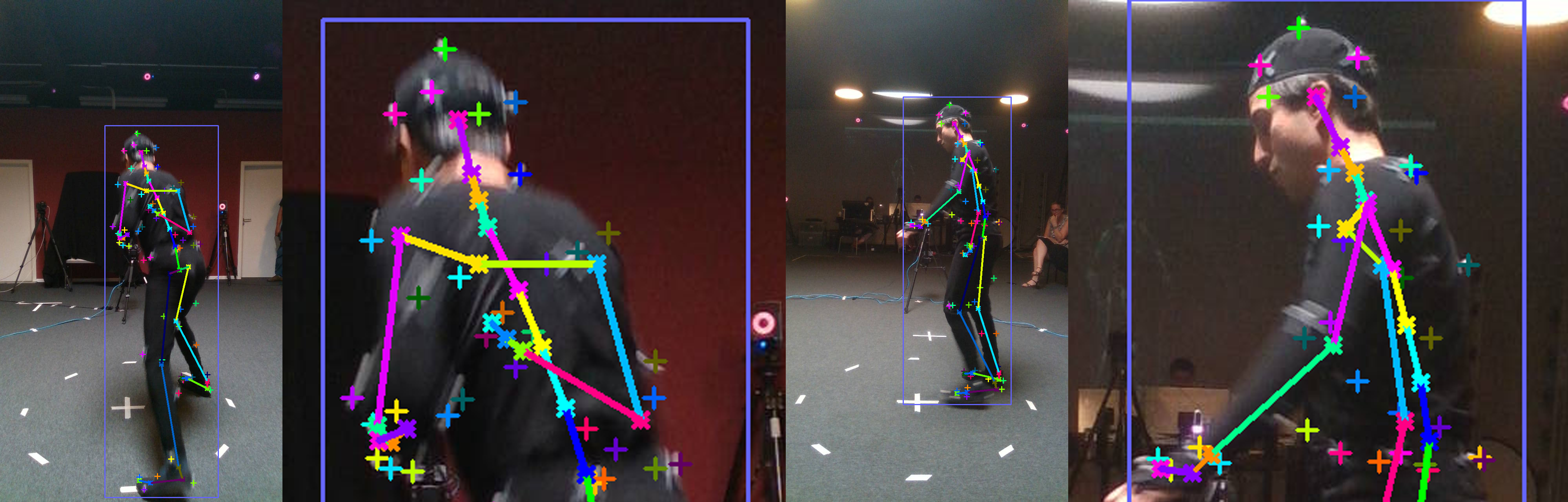}
\caption{Joint and marker 3D positions projected on color views with high accuracy. 
The marker projection accuracy which is clearly visible on the color views showcases the precision of the spatio-temporal alignment between the 3D poses (MoCap) and the RGBD data.}
\label{fig::detailed_poses}
\end{center}
\end{figure*}

\begin{figure*}
\includegraphics[width=\textwidth]{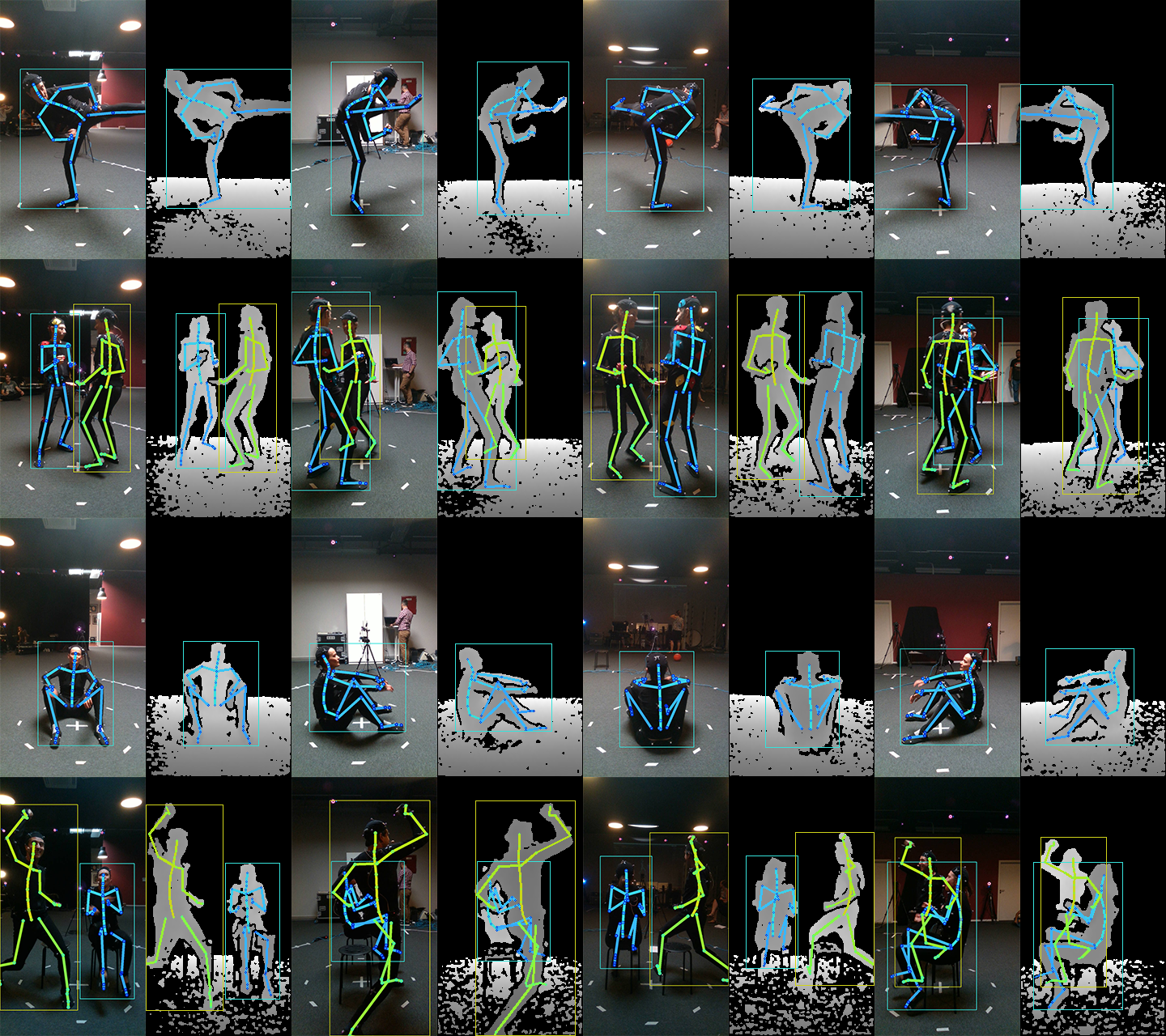}
\caption{2D pose and bounding box annotations illustrated on various color and depth frames.
The rows depict the 4 different views of mRGBD frames both from single- and two-person activities.}
\label{fig::2dannot}
\end{figure*}

\begin{figure*}[!htb]
\begin{center}
\includegraphics[width=\textwidth]{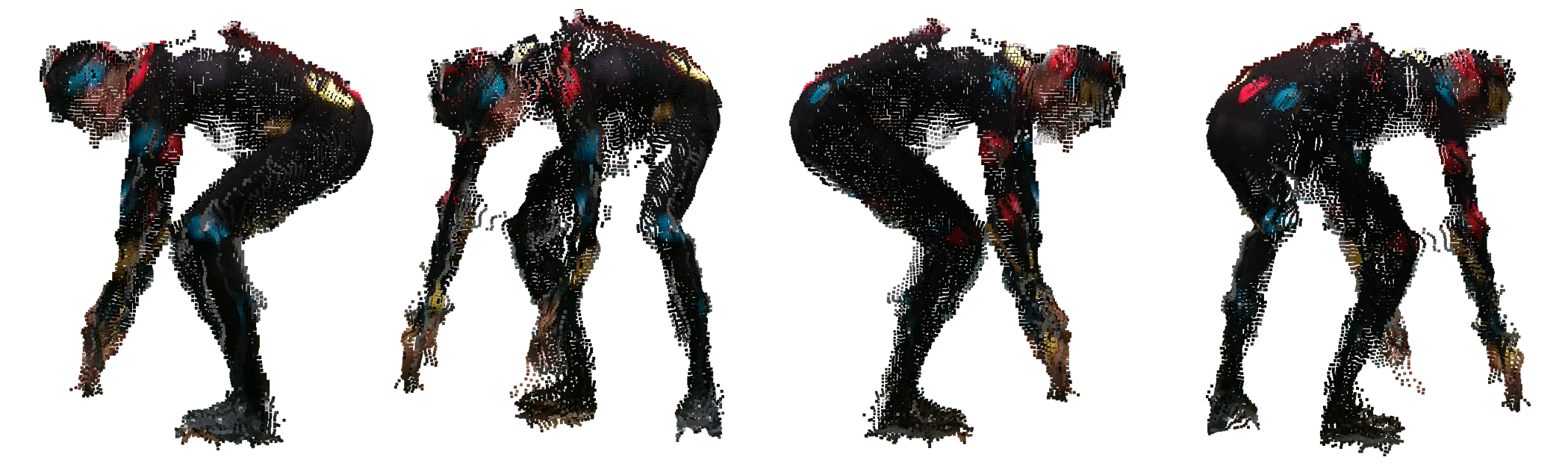}
\caption{Merged reconstructed point-cloud from one single mRGBD frame from various views.}
\label{fig::pcloud}
\end{center}
\end{figure*}

\begin{figure*}[h]
\begin{center}
\includegraphics[width=\textwidth]{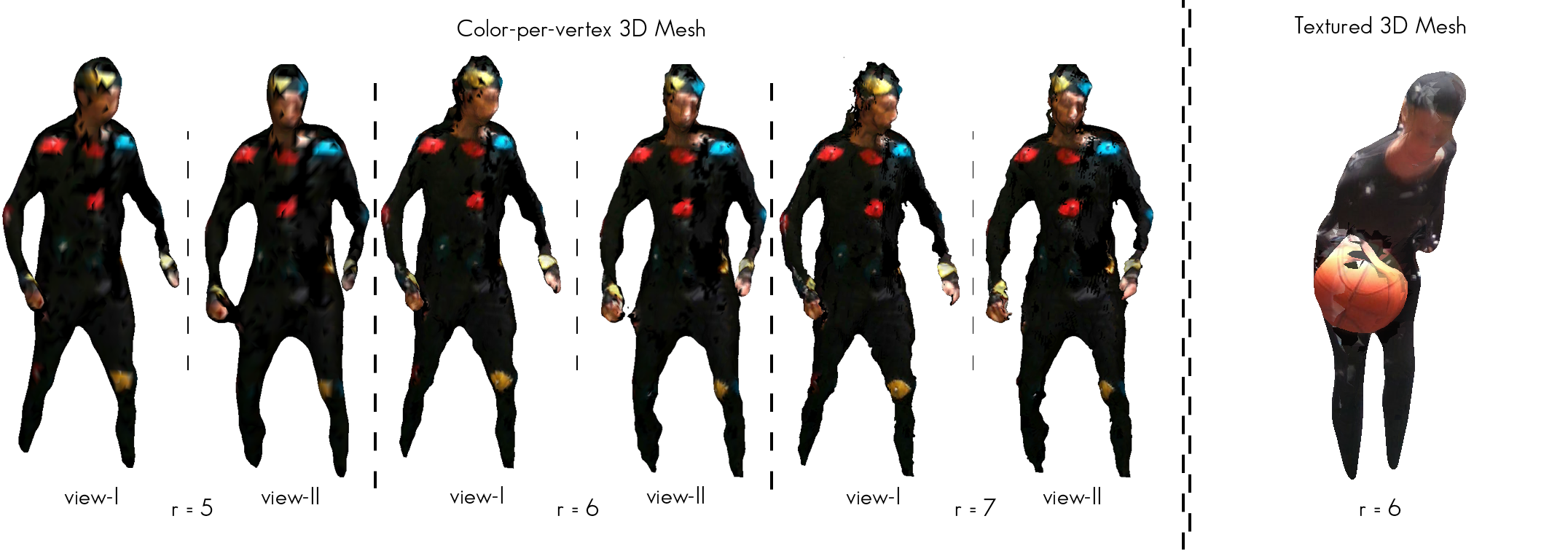}
\caption{Reconstructed \cite{alexiadis2016integrated} mesh-based volumetric data with (Left) color per vertex visualization in 3 voxel-grid resolutions, i.e. $r=5$, $r=6$ and $r=7$ and (Right) textured 3D mesh sample in voxel-grid resolution for $r=6$.}
\label{fig::meshreco}
\end{center}
\end{figure*}

\subsubsection{SYNCHRONIZATION AND CALIBRATION}\label{sec::dataset::sync_and_calib}

Inter- and intra-modality synchronization is a prerequisite for such datasets. 
The motion capture cameras operate in inter-camera synchronization by default.
With respect to the mRGBD capturing setting, as we already mentioned, Intel RealSense D415 sensors offer intra- and inter-sensor HW synchronization as well. 
With respect to the inter-modality synchronization, considering the motion capture clock as reference for the full system, along with the mRGBD and audio data timestamping, a SW-based synchronization technique was applied to temporally align the data.
In particular, given the motion capture frequency equal to $120$ Hz, the temporally closest MoCap sample to every mRGBD frame timestamp was considered the matching pose, giving a low temporal difference $t_d$, where $t_d \leq \frac{1}{120} / 2$ ms $\implies t_d \leq 4.16$ ms.  
The initial temporal offset between the modalities was detected with the use of a marker-equipped (2 markers) clapperboard at the beginning of each sequence, enabling all the modalities to capture the time instance of the clapping event.
In detail, for the motion capture data sequences, the 3D position signals of the clapperboard markers were analyzed to detect the clap event by identifying the time instance when the euclidean distance between the markers is the minimum; for the audio signals, the clap event caused an easily detectable peak on the amplitude of the audio signals, while for the RGBD data, the event was manually detected.

For the spatial alignment of the modalities, the MoCap system was calibrated once before the captures, while the mRGBD system was calibrated per subject (every subject performed all the actions at once).
The spatial alignment between MoCap and mRGBD was achieved by applying a semi-automatic technique, capturing short sequences of moving retro-reflective markers using both modalities before the capturing of each subject.
For these sequences, the infrared (IR) stream of the sensors was enabled instead of the color.
The details of the inter-modality spatial calibration go beyond the scope of this paper.

\subsubsection{2D AND 3D POSE FROM MOTION CAPTURE}
The spatio-temporal alignment between the modalities and the highly frequent and precise 3D motion capture enable the extraction of 3D poses accurately mapped on the RGBD data cues.
With a set of $J = 33$ $j$-joints, as depicted in Fig. \ref{fig::actor_rig_retarget}, a 3D pose per frame $f$ and skeleton $s$ is mapped to every single mRGBD frame.
Then, by applying inverse transformation per camera pose and projecting the 3D positions of the joints on the RGBD views, the 2D keypoints $\mathcal{K}$ are calculated by:

\begin{equation}
\label{eq:projecting}
    \mathcal{K}(f, s, j) =  \pi(\T_{\fromto{g}{l}}(x_{f,s,j}), \K_s),
\end{equation}
where  $x_{f,s, j} \in \R^3$ is the 3D position of joint $j$, $\T_{\fromto{g}{l}}$ is the transformation from the global ($g$) coordinate system to the local ($l$) one of sensor $s$ with the arrow showing the direction of the transformation. 
$\pi$ denotes the projection function that transforms the 3D coordinates to pixels, using sensor's intrinsic parameters matrix $\mathbf{\K_s}$.
The 2D outcomes of this processing are depicted in Fig. \ref{fig::detailed_poses} and \ref{fig::2dannot} .

Furthermore, considering the MoCap marker 3D positions and their corresponding 2D projections on the sensor views (using the projection of Eq. \eqref{eq:projecting}), we extract the 3D and 2D bounding boxes containing each subject per frame, by fitting a rectangular slightly padded ($2\%$ of the dimension size per side) prism and box around the 3D positions and 2D projections, respectively.

\subsubsection{VOLUMETRIC DATA FROM MULTI-VIEW RGBD}\label{sec::proc::vol}

Real-time 4D reconstruction evolves as a cutting-edge component in XR applications and beyond, especially focused on challenging dynamic data such as rigid and non-rigid human motions.
Key concept of this dataset is the exploitation of the mRGBD cues of human activities to produce and dispose volumetric data captured in a real-time manner, in the form of colored point-cloud and colored/textured 3D mesh instances for every single mRGBD frame.

\noindent \textbf{Point-cloud:} 
An RGBD image is composed of a color image $\mathcal{I}$ and a depth image $\mathcal{D}$, which, after the application of a local transformation between them, are registered to the same coordinate frame.
Then, given the depth sensors poses ($\T_v:= \left[ \begin{smallmatrix} \mathbf{R}_v&\mathbf{t}_v\\ \mathbf{0}&1 \end{smallmatrix} \right]$) known in a common coordinate system, where $\mathbf{R}_s$ and $\mathbf{t}_s$ denote rotation and translation,  respectively, we transform every depth pixel $\pixel$, $\pixel \in \mathcal{D}_s$, from the depth image domain coordinates of each view to a global coordinate system by:
\begin{equation}
\label{eq:deprojecting}
    \mathcal{T}_{\fromto{l}{g}}(\pixel) =  \T_{\fromto{l}{g}}\pi^{-1}(D_s(\pixel), \K_s, \pixel),
\end{equation}
where $\T_{\fromto{l}{g}}$ is the relative pose from the local ($l$) coordinate system of sensor $s$ to the global ($g$) one with the arrow showing the direction of the transformation. 
$\pi^{-1}$ denotes the deprojection function that transforms the pixel to 3D coordinates, using sensor's intrinsic parameters matrix $\mathbf{\K_s}$. 
Merging the transformed partial point clouds from each view to the global space, results in the colored point cloud data.
The outcome of this process is illustrated in Fig. \ref{fig::pcloud}.

\begin{figure*}[t]
\begin{center}
\includegraphics[width=\textwidth]{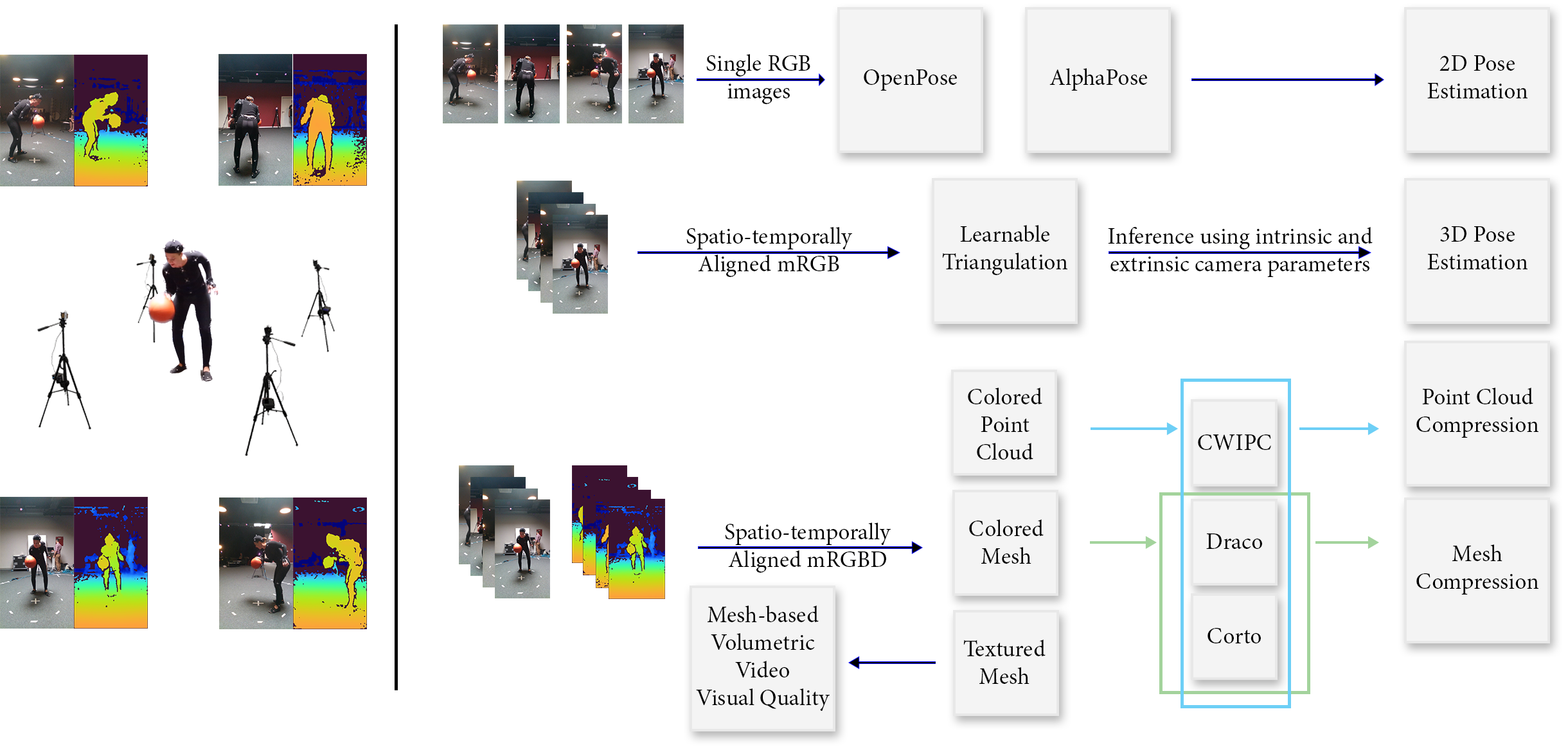}
\caption{Overview of the benchmarking schema, given the spatio-temporally aligned mRGBD frames and ground-truth poses.
Single-view RGB images are fed for 2D pose estimation.
Multi-view RGB data are used for multi-view 3D pose estimation.
Multi-RGBD frames are processed to produce point- and mesh-based volumetric video for 3D compression and visual quality benchmarking.}
\label{fig::bench_overview}
\end{center}
\end{figure*}

\noindent \textbf{3D Mesh:} 
Beyond point-based volumetric data, watertight colored and textured 3D mesh instances are reconstructed in a real-time manner (up to the frequency of the sensor acquisition, i.e. 30 fps) applying the GPU-based implementation proposed by Alexiadis \textit{et al.} \cite{alexiadis2016integrated}, based on the fast Fourier Transform (FFT) -based approach proposed by Kazhdan \cite{kazhdan2005reconstruction}.
The 3D geometry reconstruction relies on a scalar volume function $V(q)$ containing the splatted 3D surface information, as given by the point cloud calculated using the depth maps, defined over a 3D grid $q = [q_X; q_Y ; q_Z]_T \in \{1, ..., N_X\} \times \{1, ..., N_Y\} \times \{1, ..., N_Z\}$, inside the foreground object’s bounding box.
This 3D grid of $V(q)$ is considered the volume resolution of the 3D reconstruction, used with power of $2$ components for FFT, i.e. $2^r \times 2^{r+1} \times 2^r, r \in \mathbb{N}$.
Applying then the marching cubes algorithm \cite{lorensen1987marching}, the 3D surface is extracted in the form of triangular meshes (vertex positions, normal vectors and connectivity).
The coloring and texturing of each triangle of the surface is based on a weighted average between the cameras for which the specific part is not occluded. 
The weights estimation depends on the visibility angle between the camera and the respective area.
Applying \cite{alexiadis2016integrated} in voxel grid resolutions with $r = 5$, $r = 6$, $r = 7$, we extract textured and colored triangular 3D mesh instances for all the mRGBD frames of the dataset in three (3) different resolutions. 
Color-per-vertex and textured 3D mesh instances are depicted in Fig. \ref{fig::meshreco}.

\subsection{HUMAN4D BENCHMARKING SUBSETS}
For benchmarking  on HUMAN4D, we divide the dataset into two subsets, a single- (H4D1) and a two-person one (H4D2), in order to reduce the amount of data processing, as well as to evaluate samples that represent varying human poses.
At the beginning of each sequence, the subjects were standing in T-Pose for calibration purposes.
To that end, we decided to remove the first 100 frames of each sequence to avoid the collection of many similar poses (T-Pose) and to randomly sample 100 frames from the remaining part of each sequence, totaling $5600$ and $1000$ single-person and multi-person frames, respectively.
Given that we benchmark HUMAN4D with pre-trained models or non data-driven encoders, both subsets, H4D1 and H4D2, are used as testing sets.
The rest of the data can be considered as training and validation sets to allow the experimentation and development of new data-driven approaches on HUMAN4D.
We benchmark HUMAN4D with respect to pose estimation and volumetric video compression by applying state-of-the-art approaches of the respective fields.
In the following sections (Sec. \ref{sec::bench::MoCap} and \ref{sec::bench::volvideo}), we evaluate pre-trained models as well as 3D codecs for pose estimation and 3D compression respectively, on the benchmarking subsets of the dataset.
An overview of the benchmarking flow and methodology we follow and present in the following sections is depicted in Fig. \ref{fig::bench_overview}.

%%%%%%%%%%%%%%%%%%%%%%%%%%%%%%%%%%%%%%%%%%

%\section{Evaluation}\label{sec::bench}

\section{Pose estimation}\label{sec::bench::MoCap}

\begin{table*}[h]
\centering
\caption{2D pose estimation results of OpenPose \cite{cao2018openpose} and AlphaPose \cite{fang2017rmpe} with AP\textsubscript{PCKh-0.5}.}\label{tbl::2d_pose_results}
\begin{tabular}{ccccc}
\toprule 
\textit{\textbf{mAP (\%)}} & \textbf{MPII \cite{andriluka14cvpr}} & \textbf{COCO \cite{lin2014microsoft}} & \textbf{H4D1} & \textbf{H4D2}\\
 \midrule
\midrule

Cao \textit{et al.} \textsubscript{OpenPose} \cite{cao2018openpose} & 72.50 & 64.20 &	\textbf{70.02} & \textbf{68.48}
\\
Fang \textit{et al.} \textsubscript{AlphaPose} \cite{fang2017rmpe} & 82.10 &	71.00 &	\textbf{82.95} & \textbf{73.94}
\\
\midrule
\bottomrule
\end{tabular}
\end{table*}

HUMAN4D enables research to human pose-related computer vision tasks by providing spatio-temporally aligned RGBD data from multiple views under a HW-SYNC setting, along with accurate 3D and 2D poses.
Recent research efforts are devoted on various single- and multi-person pose estimation approaches, from single RGB in the wild \cite{newell2016stacked, belagiannis2017recurrent, guler2019holopose, li2018bottom}, depth \cite{xiong2019a2j, szczuko2019deep}, multi-view RGB \cite{iskakov2019learnable, kocabas2019self} and multi-view RGBD \cite{kadkhodamohammadi2017multi, carraro2018real}, among others.
However, the selection criteria of the methods we benchmark are to be open-source and applicable to HUMAN4D, producing baseline results for our dataset.
Finally, it is worth noting that the mRGBD frames of the evaluation set that go beyond the capabilities of the pre-trained models (for instance, several body parts out of at least one of the views) are excluded, preventing wrong and unfair evaluation with respect to the effectiveness of the methods.

\subsection{SINGLE-VIEW 2D POSE ESTIMATION}\label{sec::bench::single2d}
Considering the 2D poses per view, we assess state-of-the-art methods for 2D pose estimation from color images.
We apply the methods on the color views of all (4) RGBD cameras, extracting the overall error metrics per mRGB frame by averaging the errors per view.

\noindent \textbf{Methods.}
We select 2 widely known 2D pose estimation methods, a bottom-up and a top-down one, to assess their effectiveness on HUMAN4D color images.
Firstly, we select OpenPose by \textit{Cao et al.} \cite{cao2018openpose}, a deep bottom-up pose estimation method that combines confidence maps with part affinity fields to predict multi-person 2D poses in real-time.
For the evaluation of HUMAN4D, we used the latest version of the method as found to the official code repository\footnote{\url{https://github.com/CMU-Perceptual-Computing-Lab/openpose/tree/b5bffe18a8021f5f3ed98f19441b658647d9a8c3}}. 
Secondly, we evaluate AlphaPose, another data-driven approach proposed by Fang \textit{et al.} \cite{fang2017rmpe}.
AlphaPose constitutes a top-down, real-time 2D pose estimation method, that is continuously supported and updated over the last years.
For the present experiments, we used the latest version of the method as found on the official repository of the authors\footnote{\url{https://github.com/MVIG-SJTU/AlphaPose/tree/a22d3d6047b05be6ed94567c520d2a20d28d0407}}.

Finally, we also experimented with the official code of VNect\footnote{\url{http://gvv.mpi-inf.mpg.de/projects/VNect}}, by Mehta \textit{et al.} \cite{mehta2017vnect}, one of the first data-driven methods that approached 3D pose estimation from single RGB images, and A2j\footnote{\url{https://github.com/zhangboshen/A2J/tree/60b45312c5009b2053d014510c08806c2c91e950}}, by Xiong \textit{et al.} \cite{xiong2019a2j}, for 3D pose estimation from single depth maps.
However, the methods were not favorably applicable to our dataset, probably due to the differences between the characteristics of the training sets used to train the models and HUMAN4D.
For A2j for instance, the depth data used to train the body pose estimation model have been captured with Asus Xtion PRO, a structured-light depth sensor that provides depth maps of different resolution and depth noise in comparison with the stereo-based depth sensor from Intel, Intel RealSense D415.
To this end, the results are not presentable, however the related tools for experimentation are available in the code repository of our dataset\footnote{\url{https://github.com/tofis/human4d_dataset}}. 

\noindent \textbf{Metrics.} 
To measure the body joints localization accuracy, we measure mean Average Precision (mAP) for the common joints between the 2 methods and the ground truth annotations considering the Percentage of Correct Keypoints-head (\textit{PCKh}) metric, as defined in \cite{andriluka20142d}.
\textit{PCKh} constitutes a slight modification of Percentage of Correct Keypoints (\textit{PCK}) \cite{yang2012articulated}, defining a matching threshold $\alpha$ as the percentage of the head segment length (from neck to head top), instead of the long edge of the bounding box that contains the subject, aiming to make the metric independent from specific body posture and articulation.
To this end, a prediction for a frame $f$ and a skeleton $s$ is considered correct if its euclidean 2D distance error $\epsilon_{f,s}$ falls within a pixel circular region around the ground-truth keypoint with radius $r = \alpha L_{head}$, i.e.:

\begin{equation}
PCKh(f, s, j)= \left\{
\begin{array}{ll}
      1, & \epsilon_{f,s}(j) \leq  \alpha L_{head} \\
      0, & \epsilon_{f,s}(j) >  \alpha L_{head} \\
\end{array} 
\right. 
\end{equation}

\begin{equation}
AP_{PCKh}(f,\textbf{s}) = \frac{1}{\mathcal{J}_{s}}\sum_{j=1}^{\mathcal{J}_{s}}PCKh(f,s,j)
\end{equation}
where $L_{head}$ is the length of the head segment and $\alpha$ is the scalar that controls the relative threshold for correctness consideration.

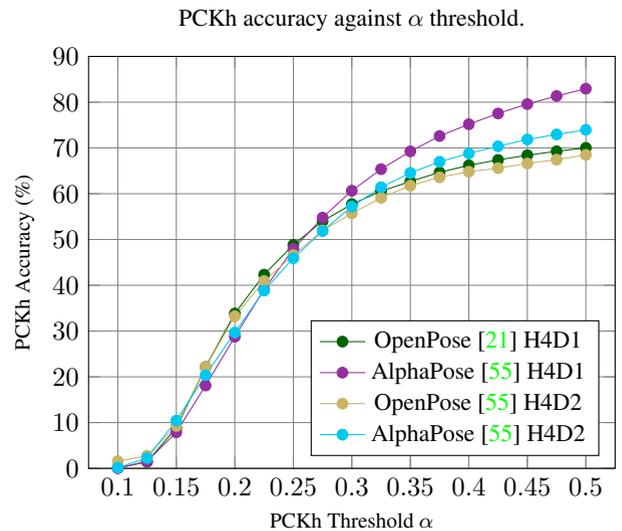
\begin{figure}[h]
\centering
\begin{tikzpicture}
\pgfplotsset{width=\columnwidth,height=7cm,compat=1.9}
\centering
\begin{axis}[
    title={PCKh accuracy against $\alpha$ threshold.},
    xlabel={PCKh Threshold $\alpha$},
    ylabel={PCKh Accuracy (\%)},
    xmin=0.075, xmax=0.525,
    ymin=0, ymax=90,
    xtick={0.1, 0.15, 0.2, 0.25, 0.3, 0.35, 0.4, 0.45, 0.5},
    ytick={0, 10, 20, 30, 40, 50, 60, 70, 80, 90},
    legend pos=south east,
    ymajorgrids=true,
    xmajorgrids=true,
    grid style={help lines},
    label style={font=\footnotesize},
	title style={font=\small},
	legend style={font=\small},
	style={thin}
]
\addplot[
    color=green0,
    mark=*,
    ]
    coordinates {
(0.1, 0.07) (0.125, 1.44) (0.15, 8.84) (0.175, 22.15) (0.2, 33.85) (0.225, 42.32) (0.25, 48.84) (0.275, 54.01) (0.3, 57.72) (0.325, 60.55) (0.35, 62.69) (0.375, 64.67) (0.4, 66.21) (0.425, 67.39) (0.45, 68.41) (0.475, 69.27) (0.5, 70.02)
    };
      \addlegendentry{OpenPose \cite{cao2018openpose} H4D1}
      
\addplot[
    color=purple0,
    mark=*,
    ]
    coordinates {
(0.1, 0.05) (0.125, 1.54) (0.15, 7.88) (0.175, 18.10) (0.2, 28.76) (0.225, 39.09) (0.25, 47.96) (0.275, 54.78) (0.3, 60.66) (0.325, 65.37) (0.35, 69.25) (0.375, 72.60) (0.4, 75.18) (0.425, 77.53) (0.45, 79.59) (0.475, 81.35) (0.5, 82.95)
    };
      \addlegendentry{AlphaPose \cite{fang2017rmpe} H4D1}
\addplot[
    color=lightgreen2,
    mark=*,
    ]
    coordinates {
(0.1, 1.564924797) (0.125, 2.756883121) (0.15, 9.264914408) (0.175, 22.10969639) (0.2, 33.2254396) (0.225, 40.9223269) (0.25, 46.59066378) (0.275, 51.97977958) (0.3, 55.71409997) (0.325, 59.10190038) (0.35, 61.75436719) (0.375, 63.5934411) (0.4, 64.80860393) (0.425, 65.59048755) (0.45, 66.60485875) (0.475, 67.44664424) (0.5, 68.48727853)
    };
      \addlegendentry{OpenPose \cite{fang2017rmpe} H4D2}
      
\addplot[
    color=cyan2,
    mark=*,
    ]
    coordinates {
(0.1, 0.164995106) (0.125, 2.250949637) (0.15, 10.43296464) (0.175, 20.3785644) (0.2, 29.67487753) (0.225, 38.84867422) (0.25, 45.90943926) (0.275, 51.8523937) (0.3, 57.18444446) (0.325, 61.42453238) (0.35, 64.54470742) (0.375, 66.99130699) (0.4, 68.79559555) (0.425, 70.35747883) (0.45, 71.85140148) (0.475, 72.95170457) (0.5, 73.94987701)
    };
      \addlegendentry{AlphaPose \cite{fang2017rmpe} H4D2}
\end{axis}
\end{tikzpicture}
\caption{OpenPose \cite{cao2018openpose} and AlphaPose \cite{fang2017rmpe} applied on the 4 views of the RGBD cameras on H4D1 and H4D2, extracting the overall error metrics per mRGB frame by averaging the errors per joint.}\label{fig::plot2d}
\end{figure}

\begin{table*}[!htb]
\centering
\caption{Single-person pose estimation results on H4D1 and CMU \cite{joo2015panoptic}.}\label{tbl::3d_pose_results}
\begin{tabular}{c|c|ccccc}
\toprule 
\textit{\textbf{Datasets}} & \textbf{CMU} & \multicolumn{4}{c}{\textbf{HUMAN4D (H4D1)}}  \\
 \midrule
 \textit{\textbf{Metrics}} & \textbf{MPJP} (cm) & \textbf{MPJP} (cm) & \textbf{RMSPJP} (cm) & \textbf{mAP (PCK\textsubscript{$\alpha_{3D}=10cm$})} & \textbf{mAP (PCK\textsubscript{$\alpha_{3D}=12.5cm$})}  \\
\midrule

Iskakov \textit{et. al} \textsubscript{LT (alg.)} \cite{iskakov2019learnable} & 2.13 & 8.42 &	9.56 & 80.26\% & 86.52\%
\\

\midrule
\bottomrule
\end{tabular}
\end{table*}

\begin{figure*}[t]
\begin{center}
\includegraphics[width=\textwidth]{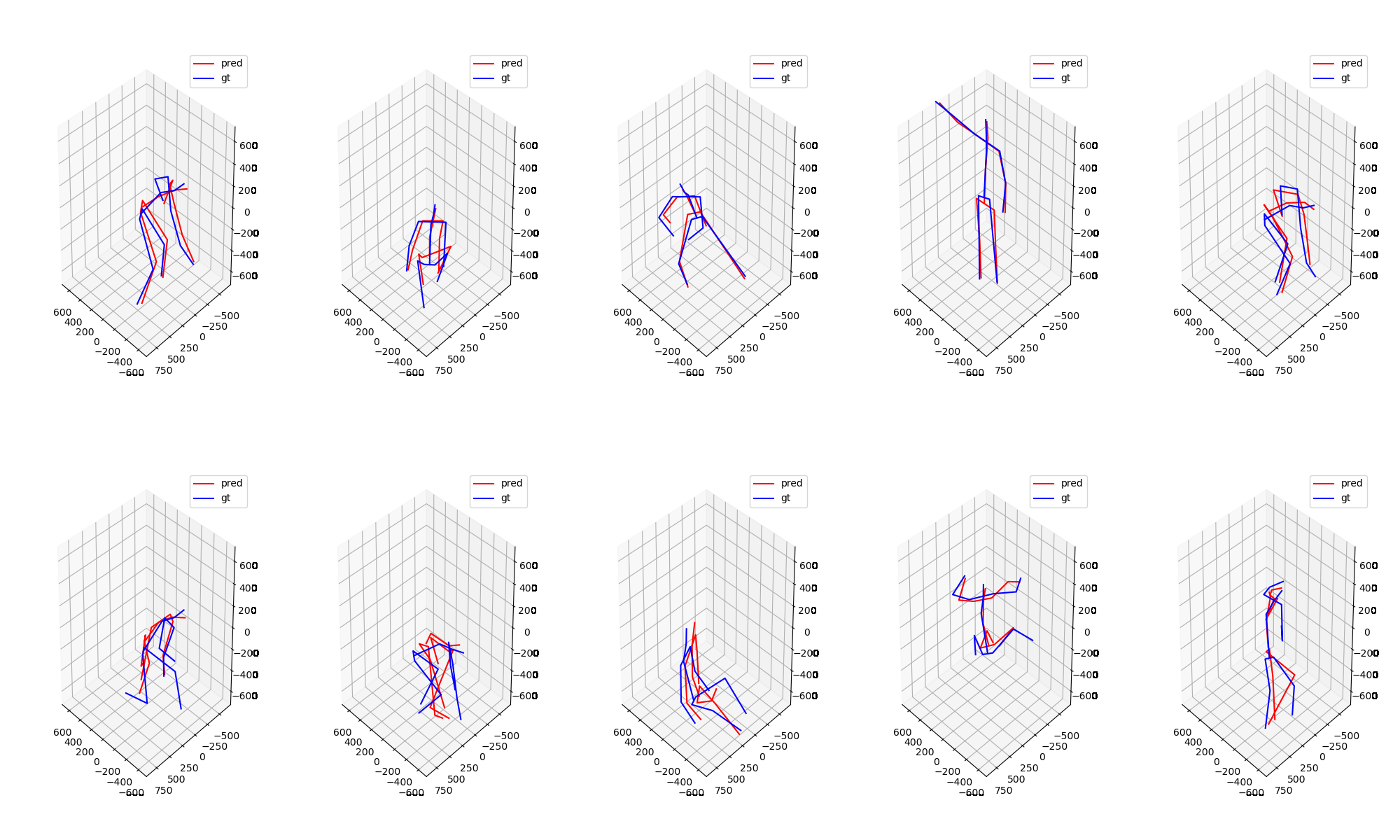}
\caption{Qualitative results of learnable triangulation (alg.) proposed by Iskakov \textit{et al.} \cite{iskakov2019learnable}. 
The top and bottom rows depict success and failure cases, respectively.
\color{blue}\textbf{Blue} \color{black} and \color{red} \textbf{red} \color{black} colors correspond to \color{blue} \textbf{ground truth} \color{black} and \color{red} \textbf{predicted} \color{black} poses.}
\label{fig::3d_vis}
\end{center}
\end{figure*}

\noindent \textbf{Results.}
We separately present the results of the methods on H4D1 and H4D2 to better distinguish their effectiveness on single- and multi-person color data.
At first, similarly to the outcomes on other public datasets, AlphaPose outperforms OpenPose showing higher accuracy both in single- and multi-person benchmarking sets of HUMAND.
Nevertheless, even though both methods showcase lower accuracy on the multi-person data of H4D2, which is much more challenging due to the occlusions between the subjects, it is worth noting that the difference between the single- and multi-person results of OpenPose is low ($\sim1.5\%$), while AlphaPose presents a higher drop of approximately $9\%$.
Taking into account that the distance between the subjects and the sensors is short, from 1 to 2 meters, and in most of the two-person samples, there are severe occlusions for some of the sensors, we can probably assume that OpenPose, as a bottom-up approach behaves more robustly on occlusions, however AlphaPose, as a top-down approach, is more accurate but is strongly affected by occlusions.
In order to provide extra information to the reader, along with the results on HUMAN4D, we also indicate the related outcomes of the methods to other datasets, i.e. MPII \cite{andriluka14cvpr} and COCO \cite{lin2014microsoft} using \textit{PCKh} with $\alpha = 0.5$, as presented in Table \ref{tbl::2d_pose_results}. 
Finally, a plot depicting the correlation between \textit{PCKh} mAP against $\alpha$ threshold for both methods on both subsets, is illustrated in Fig. \ref{fig::plot2d}.

\subsection{MULTI-VIEW 3D POSE ESTIMATION}\label{sec::bench::MoCap_multi}

Subsequently, we evaluate multi-view 3D pose estimation on HUMAN4D, exploiting the multi-view color images along with the respective intrinsic and extrinsic camera parameters and using HUMAN4D 3D poses as ground truth. 

\noindent \textbf{Methods.} 
We choose a recent state-of-the-art method proposed by Iskakov \textit{et al.} \cite{iskakov2019learnable}, which constitutes a novel solution for multi-view single-person 3D human pose estimation based on a learnable triangulation (LT) technique, combining 3D information from multiple spatio-temporally aligned 2D color views.
In particular, LT\textsubscript{(alg.)} \cite{iskakov2019learnable} is a top-down 3D pose estimation method based on end-to-end differentiable algebraic triangulation with an addition of confidence weights estimated from the input images.
We ran the experiments only on the HD41 benchmarking subset of the dataset since the method estimates single-person 3D poses, using the latest version of the code published by the authors\footnote{\url{https://github.com/karfly/learnable-triangulation-pytorch}}.

\noindent \textbf{Metrics.}
With respect to the metrics, we use the Mean Per Joint Position (MPJP) \cite{mehta2017vnect} and Root Mean Squared Per Joint Position (RMSPJP) error metrics, which both are influenced by large outliers, however the latter better incorporates the variance of the estimates and their bias. 
For a frame $f$ and a skeleton $s$, MPJP and RMSPJP are computed as:
\begin{equation}
\epsilon_{f, s}(j) = ||\hat{x}_{f, s}(j) - x_{f, s}(j))||_{2}
\end{equation}

\begin{equation}
\mathcal{E}_{MPJP}(f,s) = \frac{1}{\mathcal{J}_{s}}\sum_{j=1}^{\mathcal{J}_{s}}\epsilon_{f, s}(j)
\end{equation}

\begin{equation}
\mathcal{E}_{RMSPJP}(f,\textbf{s}) = \sqrt{\frac{1}{\mathcal{J}_{s}}\sum_{j=1}^{\mathcal{J}_{s}}\epsilon^2_{f, s}(j)}
\end{equation}
where $\mathcal{J}_{s}$ is the total number of joints of skeleton $s$. 
Finally, we also use mean AP with 3D PCK metric \cite{mehta2017monocular} per joint, where an estimate is considered correct when the 3D euclidean distance error, i.e. $\epsilon_{f, s}(j)$, is less than a distance threshold $\alpha_{3D}$, as:

\begin{equation}
PCK_{3D}(f,s,j)= \left\{
\begin{array}{ll}
      1, & \epsilon_{f, s}(j) \leq \alpha_{3D} \\
      0, & \epsilon_{f, s}(j) > \alpha_{3D} \\
\end{array} 
\right. 
\end{equation}

\begin{equation}
AP_{PCK_{3D}}(f,s) = \frac{1}{\mathcal{J}_{s}}\sum_{j=1}^{\mathcal{J}_{s}}PCK_{3D}(f,s,j)
\end{equation}
for a frame $f$ and skeleton $s$, correspondingly.

\begin{figure}[h]
\centering
\begin{tikzpicture}
\pgfplotsset{width=\columnwidth,height=5cm,compat=1.9}
\centering
\begin{axis}[
    title={mAP using 3D PCK against $\alpha_{3D}$ threshold in $cm$.},
    xlabel={3D PCK Threshold $\alpha_{3D}$ ($cm$)},
    ylabel={mAP (\%)},
    xmin=4.5, xmax=20.5,
    ymin=0, ymax=100,
    xtick={5, 7.5, 10, 12.5, 15, 17.5, 20},
    ytick={25,50,75,100},
    legend pos=south east,
    ymajorgrids=true,
    xmajorgrids=true,
    grid style={help lines},
    label style={font=\footnotesize},
	title style={font=\small},
	legend style={font=\small},
	style={thin}
]
\addplot[
    color=lightgreen2,
    mark=*,
    ]
    coordinates {
(5.0, 6.44) (6.25, 39.61) (7.5, 62.12) (8.75, 73.05) (10.0, 80.26) (11.25, 83.67) (12.5, 86.52) (13.75, 88.48) (15.0, 90.32) (16.25, 92.34) (17.5, 93.76) (18.75, 94.83) (20.0, 95.79)
    };
      \addlegendentry{Iskakov \textit{et al.} \textsubscript{LT (alg.) \cite{iskakov2019learnable}} }  
\end{axis}
\end{tikzpicture}
\caption{Benchmarking of Algebraic Learnable Triangulation \cite{iskakov2019learnable} on H4D1 using total 3D PCK results in different $\alpha_{3D}$ threshold values in $cm$.}\label{fig::qual3d}
\end{figure}
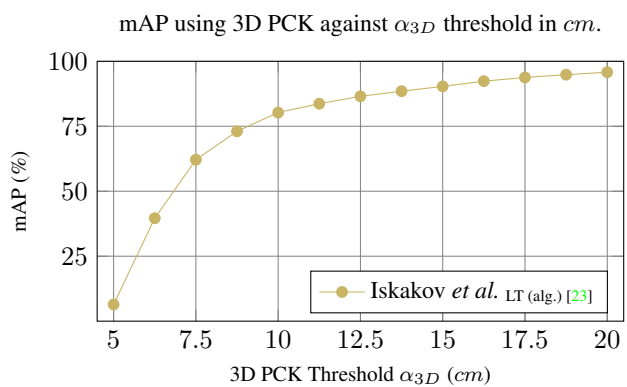

\noindent \textbf{Results.}
Classic triangulation algorithms assume that the 2D point coordinates from each view equally contribute to the triangulation 3D point coordinates estimation.
The major advantage of the LT approach is that the contribution of the 2D joint positions that cannot be estimated reliably (e.g. due to joint occlusions) to the final triangulation outcome,  is controlled by a neural network.
In particular, learnable weights have been added to the coefficients of the matrix corresponding to different views.
A limitation of the LT approach is that it fails when some of the body parts are out of the field of view of the cameras, leading to erroneous estimates.
Another limitation is that LT approach supports only single-person 3D pose estimation and for that reason it was applied only on H4D1.
Quantitative results of the method on HUMAN4D, complemented with results on CMU \cite{joo2015panoptic} dataset, are reported in Table \ref{tbl::3d_pose_results}.
Fig. \ref{fig::qual3d} illustrates the correlation between the mAP against $\alpha_{3D}$ threshold on HUMAN4D.
Qualitative results regarding the predicted 3D poses against ground-truth on HUMAN4D are illustrated in Fig. \ref{fig::3d_vis}, where LT\textsubscript{(alg.)} seems accurate in "clean" poses where self-occlusions are limited (success cases on top rows), while the accuracy is limited in the presence of self-occlusions (failure cases on bottom rows).

\section{Volumetric Video}\label{sec::bench::volvideo}
Beyond pose estimation, we benchmark a set of state-of-the-art static 3D codecs, in the context of a live streaming scenario.
Moreover, we assess the visual quality of textured 3D mesh instances to demonstrate the positive correlation between the objective visual quality and the FFT voxel-grid resolution.

\subsection{Volumetric Video Compression}\label{sec::bench::volvideo_comp}

Compression of volumetric data produced in a real-time manner is thought to be a key enabler of a wide variety of applications, such as XR teleconference, real-time dense surface mapping in AR devices and free-viewpoint videos.
A key contribution of HUMAN4D is that it enables future benchmarking in static and temporal volumetric video compression, by offering a large dataset of samples and sequences of point- and mesh-based volumetric data.
In contrast with motion pictures where solutions are mature and proven, real-time varying geometry coding is still an open challenge frequently cured utilizing only intra-frame coding, ignoring temporal relations between volumes of consecutive frames.
Such an endeavour is presented in \cite{doumanoglou2019benchmarking} by Doumanoglou et al.
In a similar manner, for the purpose of this work, the codecs are tested in various profiles, aiming at specific bit-rates, using appropriate metrics on HUMAN4D point- and mesh-based volumetric data cues.
To be coherent, we define common codec profiles both for H4D1 and H4D2 dataset subsets.
A matching procedure between different codecs for the same target bit-rate was adopted, defining the acceptable deviation margin between target and achieved bit-rate to be $\pm$10\%.

\subsubsection{MESH-BASED VOLUMETRIC VIDEO COMPRESSION}\label{sec::bench::compression_intro}
Initially, we benchmark 3D codecs on mesh-based volumetric data using the benchmarking subsets of meshes reconstructed in three different voxel-grid resolutions (i.e. $r=\{5, 6, 7\}$) applying the real-time 3D reconstruction method by Alexiadis \textit{et al.}, as reported in Section \ref{sec::proc::vol}.

\noindent \textbf{Codecs.}
We employ Corto \cite{Corto} and Draco \cite{Draco}, two 3D codecs particularly chosen due to their high quality real-time performance. 
Targeting specific bit-rates for real-time mesh-based volumetric video transmission, we constructed a series of compression profiles with varying compression level, quantization parameter per attribute and different compression methods for specific attributes. 
HUMAN4D mesh-based compression benchmarking focuses on three different per-vertex attributes: 
\textit{geometry} and \textit{normals} represented in floating points and \textit{color} in unsigned integers.

Corto codec \cite{Corto} configuration consists of four parameters. 
One quantization value for each of the mesh attributes, i.e. Geometry (GQ), Normal (NQ) and Color (CQ) Quantization bits, and one switch to denote the normal prediction method. 
We select between two different normal prediction methods, the Normals Quantized Coding (NQC) and the Normals Delta Coding (NDC).
In the former, we store the differences between the normals estimated from the quantized geometry and the quantized actual normals, using an octahedron projection representation \cite{meyer2010floating}. 
In the latter, the quantized normals in the octahedron projection representation are solely delta coded, with respect to a neighboring quantized normal belonging to a quad incident to the normal’s vertex.

Regarding the Draco codec \cite{Draco}, the configurable parameters are the compression level (CL) which adjusts the compression speed versus the size mixture, the geometry quantization bits (GQ), the normals quantization (NQ) and the color quantization bits (CQ). 
Contrary to Corto, Draco does not expose any normal manipulation option to adjust.

Beyond these conventional open-source codecs, novel 3D and 4D data compression approaches have appeared, such as the one proposed by Tang \textit{et al.} \cite{tang2020deep}.
This method constitutes a novel block-based 3D compression model, being the first deep 3D compression method that can train end-to-end with entropy coding, lossless compression of the surface topology, exhibiting a novel block-based texture parametrization that inherently promotes temporal consistency without tracking and the necessity of the UV coordinates compression.
This codec achieves superior results in comparison to conventional 3D codecs, such as Draco and Corto, in regards with the rate-distortion (RD) balance.
Specifically, it is deemed to achieve on average $66\%$ lower bit-rate for the same level of distortion in 4D data. 
For the purpose of this work, we did not benchmark this particular codec since it is not currently open-source.

\input{sections/compression_plots.tex}

\noindent \textbf{Metrics.}
With respect to the metrics, we use \textit{RMS}, \textit{HausdorffAbs} and \textit{HausdorffRel} metrics to compare the compressed and raw mesh-based representations. 
For the extraction of \textit{RMS} and \textit{Hausdorff} distance, we exploit a tool implemented based on \cite{cignoni1998metro}.
This tool provides numerical metrics for the similarity of source and target triangle or quadrilateral meshes. 
It is worth mentioning that, for the same pair, swapping between the source and target meshes can lead to different numerical values, thus as usual for these metrics in the literature, we define the correct value to be the maximum of these two, for all metrics.

%To better describe the distortion metrics, let us define the distance $d(p, S')$ between a point $p$ belonging to a surface $S$ and a surface $S'$ as:
%\begin{equation}
%\label{metrics:point_surface_distance}
%d(p, S') = \min_{p'\in S'} \norm*{p - p'}_2.
%\end{equation}
%Then the one-sided distance between the two surfaces can be defined as:
%\begin{equation}
%\label{metrics:distance_between_surfaces}
%E(S, S') = \max_{p'\in S_1} d(p, S').
%\end{equation}
%It is crucial to mention that this definition is not symmetric as there are surfaces that $E(S, S') \neq E(S', S)$.
%At this point, we calculate all the minimum distances from each point of the source surface to the target surface.
\textit{Hausdorff distance} metric is used in two variations.
\textit{HausdorffAbs metric} is defined as the maximum value of all the uniformly minimum sampled distances across all points of the source surface to the target surface. 
\textit{HausdorffRel metric} is a variation of \textit{HausdorffAbs metric} which tackles the comparison of surfaces with different scales.
For the \textit{RMS} calculation, we need to have a set of minimum distances between two surfaces, the mean distance $E_m$ can be calculated by:
\begin{equation}
\label{metrics:mean_distance_between_surfaces}
E_m(S, S') = \frac{1}{|S|} \int_{S}^{} d(p, S') dS
\end{equation}
where $|S|$ denotes the area of $S$. 
Using the mean distance formula, the \textit{root mean square error} is defined by:
\begin{equation}
\label{metrics:rms_between_surfaces}
RMS_{S\rightarrow S'} = \sqrt{\frac{1}{|S|} \int_{S}^{} d(p, S') dS}.
\end{equation}

\noindent \textbf{Results.}
For a fair comparison between the codecs, we choose to employ a testing scheme based on rate-distortion terms.
In that direction, we keep the bit-rates steady for the pairs and evaluate the corresponding distortion introduced by each codec.
As it can be seen in Fig. \ref{fig::plots_mesh},
Draco consistently outperforms Corto, in terms of distortion induced for any tested bit-rate.
The profiles used for the benchmarking are depicted in Table \ref{tab:mesh_codec_profiles}.

Having tested the same codec profiles both for single and multi-person subsets of the HUMAN4D dataset, we noticed that the bit-rates achieved by both codecs on the multi-person subset are slightly greater than those on the single-person one. 
That is probably due to the fact that the additional information induced in the form of the second subject, leads to larger surfaces that, despite using the same voxel-grid areas and resolutions, results in more challenging 3D surfaces to compress, in regards with elements count and connectivity information.

\subsubsection{POINT-BASED VOLUMETRIC VIDEO COMPRESSION}\label{sec::bench::pc}

To benchmark point cloud compression, beyond the reconstruction of the raw point-cloud instances from the mRGBD samples described in Section \ref{sec::proc::vol}, we also use another point-cloud reconstruction approach. 
The raw point-cloud instances typically contain $\sim25,000$ points per frame for the single-subject sequences and $\sim40,000$ points for the two-subject ones. 
This alternative reconstruction approach allows us to create denser point clouds by sampling points from the surface of the high resolution meshes (i.e. using voxel-grid resolution with $r=7$). 
Points are sampled from the mesh surface with a probability proportional to the area of the underlying mesh faces using Point Cloud Library (PCL) \cite{Rusu:pcl}. 
We set the algorithm to generate point cloud instances containing $300,000$ points per frame. 

\noindent \textbf{Codecs.} To benchmark the performance of point cloud compression, we perform a rate-distortion analysis for the codecs Draco, Corto and CWIPC, the MPEG anchor codec proposed in \cite{mekuria2016design} and evaluated in \cite{cwipcc:eval}. 
CWIPC is parameterizable with respect to the Octree Depth (OD) and JPEG Quantization Parameter (JPEGQP).
We select to perform the analysis on 4 target bit-rates. Note that, for all codecs we first identified the compression parameters that achieve the target bit-rates within a 10\% tolerance. 
Details on these profiles are listed in Table \ref{tbl::point_codecs}.

\noindent \textbf{Metrics.}
To measure the distortions introduced by compression to the point-cloud samples, we used standard, well established, full reference metrics, as released by the standards body MPEG \cite{mpeg:cfp, mpeg:emergingStandard}. 
More specifically, we measure Peak Signal-to-Noise Ratio (PSNR) using the maximum of the nearest neighbor euclidean distances amongst all points in the reference point cloud as the peak value $v_p$ by:
\begin{equation}\label{eq:psnr}
    PSNR = 10\log(\frac{v_p^{2}}{MSE})
\end{equation}

The same process is then applied to the point cloud colors at each of the corresponding points between the decoded and the groundtruth point clouds. 
Metrics are collected utilizing the MPEG PCC-DMETRIC tool \cite{mpegPCmetric}\footnote{\url{http://mpegx.int-evry.fr/software/MPEG/PCC/mpeg-pcc-dmetric}} to calculate these distortions for each frame in the dataset.

\noindent \textbf{Results.}
Analyzing the experimental results, CWIPC codec achieves lower geometry distortions for the same bit-rate in comparison with Draco and Corto, while in higher bit-rates, all the benchmarked codecs showcase similar efficiency.
CWIPC exploits octree occupancy to encode geometry positions, thus is able to retain more points from the original point cloud. 
Details with respect to point-cloud compression benchmarking are illustrated in Fig. \ref{fig::plots_pc}, while the codec profiles used for the experiments are listed in Table \ref{tbl::point_codecs}.
For the sake of clarity, we summarize the abbreviations of codec configuration parameters in Table \ref{tab:abbreviations}.

\begin{table}[h]
\centering
\begin{tabular}{ccc}
\toprule
\textbf{Codecs} & \textbf{Parameter} & \textbf{Abbreviation} \\ 
\midrule
Draco & Compression Level & CL \\
Draco/Corto & Normal Quantization Bits & NQ \\
Draco/Corto & Geometry Quantization Bits & GQ \\
Draco/Corto & Color Quantization Bits & CQ \\
Corto & Normals Quantized Coding & NQC \\
Corto & Normals Delta Coding & NDC \\
CWIPC & Octree Depth & OD \\
CWIPC & JPEG Quantization Parameter & JPEGQP \\
\bottomrule
\end{tabular}
\caption{Abbreviations.}
\label{tab:abbreviations}
\end{table}

\subsection{Mesh-based Volumetric Video Visual Quality}

%%%%%%%%%%%%%%%%%%%%%% PETROS %%%%%%%%%%%%%%%%%%%%%%

In this section, we assess the visual quality of HUMAN4D textured 3D mesh instances between the three different resolutions of the underlying voxel-grid.
The aim is to demonstrate the positive correlation between the objective visual quality and the utilized voxel-grid resolution used to reconstruct the mesh-based volumetric data.

As mentioned in Section \ref{sec::proc::vol}, the reconstruction of the mesh-based volumetric data is achieved by applying the real-time method proposed by Alexiadis \textit{et al.} \cite{alexiadis2016integrated}, parameterized in three different voxel-grid resolutions to produce watertight textured 3D mesh instances of varying vertex and face counts.
Higher resolution grids lead to meshes of higher element count that are, per se, expected to capture more photorealistically and precisely the observed subjects.

Apart from the self-evident impact of higher resolution sampling on the reconstructed hull's spatial fidelity, additional benefits may arise with regard to the accurate colorization of its surface.
To showcase and quantify this effect, we firstly project the examined mesh on its respective RGB images and sample the color of its fragments based on a weighted contribution of the corresponding pixels.
Then, we render the mesh from the exact same viewpoints that the aforementioned images were captured and compare the synthesized images to their respective silhouette-cropped textures, using conventional image quality metrics.

%The subset of our data-set we used for benchmarking in this chapter is divided into two parts, a single- and a multi-person one.
We conduct the assessment separately to H4D1 and H4D2 benchmarking subsets.
The former, consisting of 4 subjects with 14 sequences each, and each of these sequences with 100 sampled mRGBD frames, reconstructed in 3 voxel-grid resolutions (i.e. $r=\{5, 6, 7\}$) and rendered from 4 viewpoints, results in a total of $67,200$ rendered views of $16,800$ mesh instances.
Similarly, the latter includes 2 couples, with 5 sequences of 100 frames each, reconstructed in the same 3 voxel-grid resolutions and rendered from corresponding viewpoints, giving a total of $12,000$ views of $3,000$ 3D meshes.

\noindent \textbf{Metrics.} For the visual quality assessment, we opted to use \textit{Peak Signal-to-Noise Ratio (PSNR)} (Eq. \ref{eq:psnr}) and \textit{Structural Similarity Index (SSIM)} as metrics to objectively quantify the photometric and photorealistic consistency between the captured, raw color (RGB) views and the mesh-based 4D representations in the various voxel-grid resolutions on the rendered views' quality.

\textit{SSIM} is a full-reference metric conceived as an improvement over the traditional \textit{PSNR} and \textit{MSE-family} metrics and is widely referenced in the video and photography industry as it is believed to capture better the human perception of visual quality.
Instead of decomposing the input signals and then estimating absolute errors, as in the case of MSE-like metrics, \textit{SSIM} incorporates into its calculations the fact that images are inherently highly structured and thus their topology and the relations that arise between their elements, due to that fact, should not be ignored.
Luminance Masking and Contrast Masking are two well-known visual perception phenomena that are taken into account during the process of obtaining \textit{SSIM} measurements.
The former is about the low visibility of distortions in bright regions, while the latter is about the masking of distortions in highly textured, non-smooth, areas of an image.

The \textit{SSIM} formula is composed of three individual measurements of "structural similarity", luminance $l$, contrast $c$ and structure $s$ between two windows $x$ and $y$ of similar size.
The individual comparison formulas are:
\begin{equation}
l(x,y)=\frac{2\mu_x\mu_y + c_1}{\mu^2_x + \mu^2_y + c_1}
\end{equation}
\begin{equation}
c(x,y)=\frac{2\sigma_x\sigma_y + c_2}{\sigma^2_x + \sigma^2_y + c_2}
\end{equation}
\begin{equation}
s(x,y)=\frac{\sigma_{xy} + c_3}{\sigma_x\sigma_y + c_3}
\end{equation}
\noindent with $\mu_x$ the average of $x$, $\mu_y$ the average of $y$, $\sigma^2_x$ the variance of $x$, $\sigma^2_y$ the variance of $y$, $\sigma_{xy}$ the covariance of $x$ and $y$, $c_1=(k_1L)^2$, $c_2=(k_2L)^2$, $c_3=(c_2/2)$ are three variables to stabilize the division with weak denominator, $L$ the dynamic range of the pixel values and $k_1=0.01 , k_2=0.03$ by default.
\textit{SSIM} is then a weighted combination of these comparative measures:
\begin{equation}
SSIM(x,y)=[l(x,y)^\alpha \cdot c(x,y)^\beta \cdot s(x,y)^\gamma]
\end{equation}
\noindent where $\alpha, \beta, \gamma > 0$ are parameters used to adjust the relative importance of the three components.
More on the SSIM and its development can be found in \cite{wang2004image}.

\begin{table}[]
\centering
\caption{Single-person PSNR and SSIM}
\begin{tabular}{c|*3c|*3c}
\toprule
 & \multicolumn{3}{c}{\textbf{PSNR}} & \multicolumn{3}{c}{\textbf{SSIM}}\\
\midrule
\textit{\textbf{Subject}} & \textbf{r=5} & \textbf{r=6} & \textbf{r=7} & \textbf{r=5} & \textbf{r=6} & \textbf{r=7}\\
\midrule
\textbf{S1} & 36.18 & 36.59 & \textbf{36.70} & 0.98598 & 0.98685 & \textbf{0.98707}\\
\textbf{S2} & 34.51 & 34.84 & \textbf{34.89} & 0.98320 & 0.98389 & \textbf{0.98395}\\
\textbf{S3} & 33.48 & 33.71 & \textbf{33.73} & 0.98235 & \textbf{0.98278} & 0.98270\\
\textbf{S4} & 33.36 & 33.54 & \textbf{33.55} & 0.98262	& \textbf{0.98302} & 0.98293\\
\midrule
\textbf{Average}	& 34.38 & 34.67 & \textbf{34.72} & 0.98354 & 0.98413 & \textbf{0.98416}\\
\bottomrule
\end{tabular}
\label{fig::drak_single}
\end{table}

\begin{table}[]
\centering
\caption{Multi-person PSNR and SSIM}
\begin{tabular}{c|*3c|*3c}
\toprule
 & \multicolumn{3}{c}{\textbf{PSNR}} & \multicolumn{3}{c}{\textbf{SSIM}}\\
\midrule
\textit{\textbf{Subjects}} & \textbf{r=5} & \textbf{r=6} & \textbf{r=7} & \textbf{r=5} & \textbf{r=6} & \textbf{r=7}\\
\midrule
\textbf{S1 \& S2}	& 32.59 & 33.02 & \textbf{33.26} & 0.97513 & 0.97634 & \textbf{0.97720}\\
\textbf{S3 \& S4}	& 38.20	& 38.37	& \textbf{38.41}	& 0.98346 &	0.98432	& \textbf{0.98488}\\
\midrule
\textbf{Average}	& 35.39	& 35.70	& \textbf{35.84}	& 0.97930 & 0.98033 & \textbf{0.98104}\\
\bottomrule
\end{tabular}
\label{fig::drak_multi}
\end{table}

\begin{figure*}
\begin{subfigure}{0.25\textwidth}
  \centering
  \includegraphics[width=1.0\linewidth]{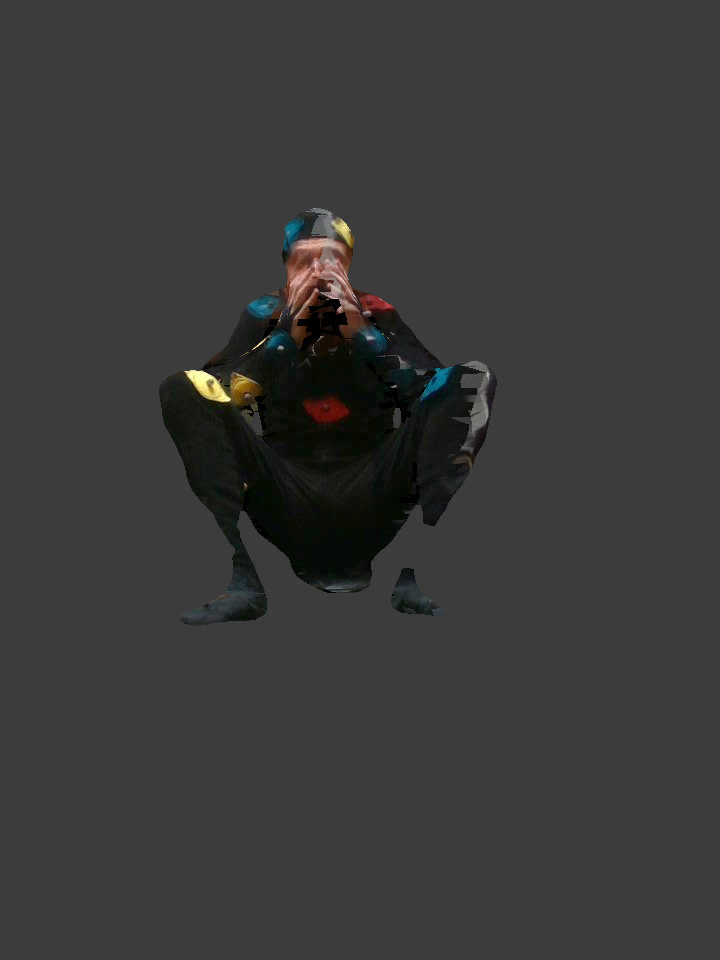}
\caption{$r=5$}
\end{subfigure}%
\begin{subfigure}{0.25\textwidth}
  \centering
  \includegraphics[width=1.0\linewidth]{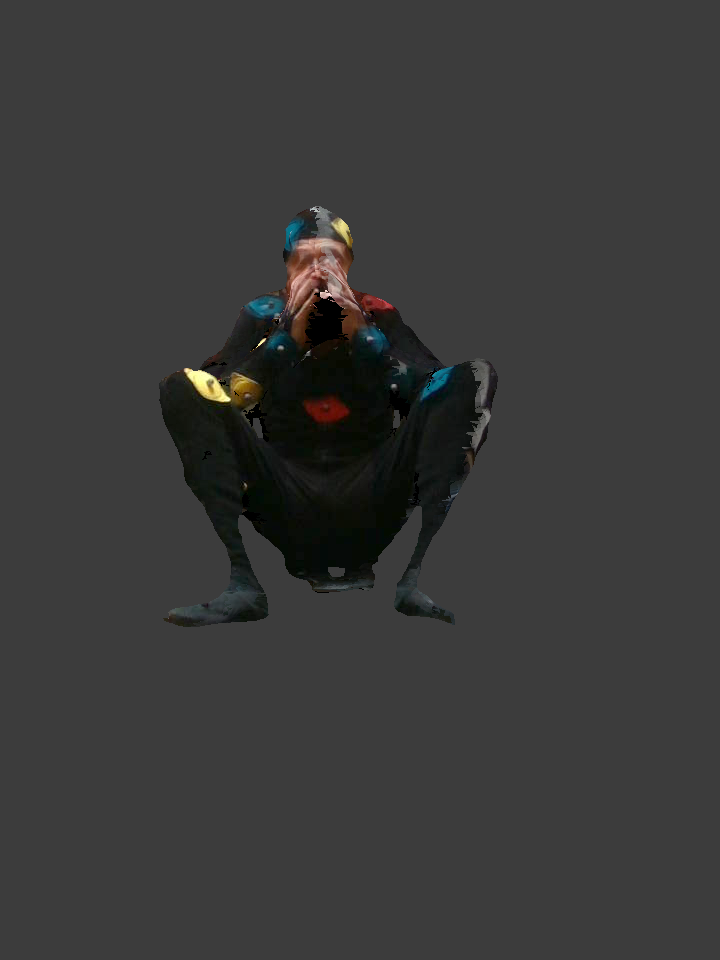}
\caption{$r=6$}
\end{subfigure}%
\begin{subfigure}{0.25\textwidth}
  \centering
  \includegraphics[width=1.0\linewidth]{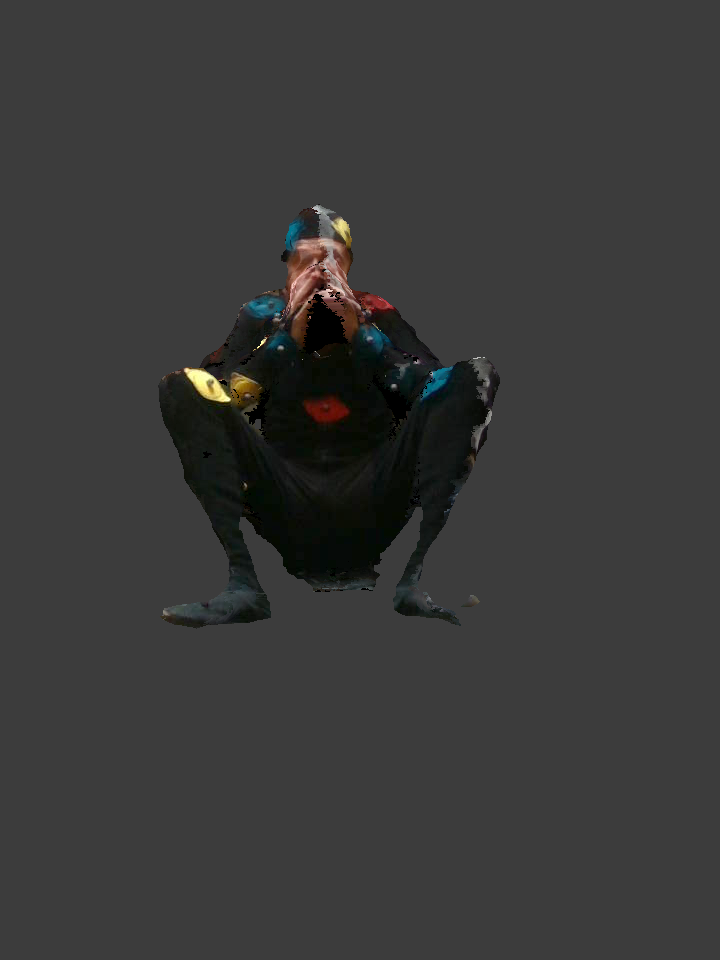}
\caption{$r=7$}
\end{subfigure}%
\begin{subfigure}{0.25\textwidth}
  \centering
  \includegraphics[width=1.0\linewidth]{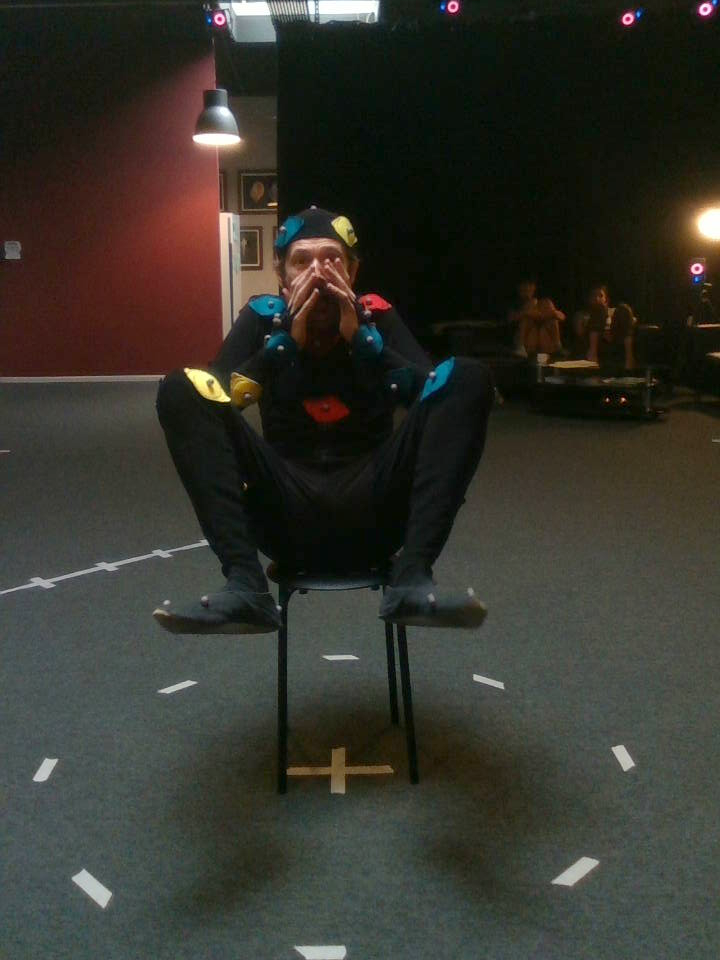}
\caption{\textit{RGB}}
\end{subfigure} \\
\begin{subfigure}{0.25\textwidth}
  \centering
  \includegraphics[width=1.0\linewidth]{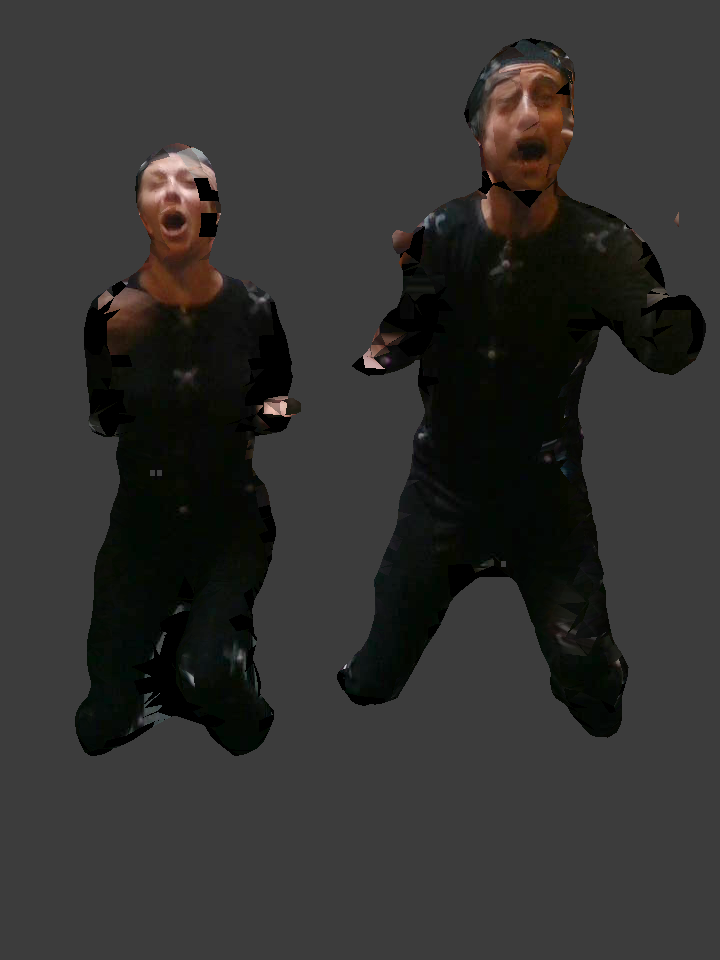}
\caption{$r=5$}
\end{subfigure}%
\begin{subfigure}{0.25\textwidth}
  \centering
  \includegraphics[width=1.0\linewidth]{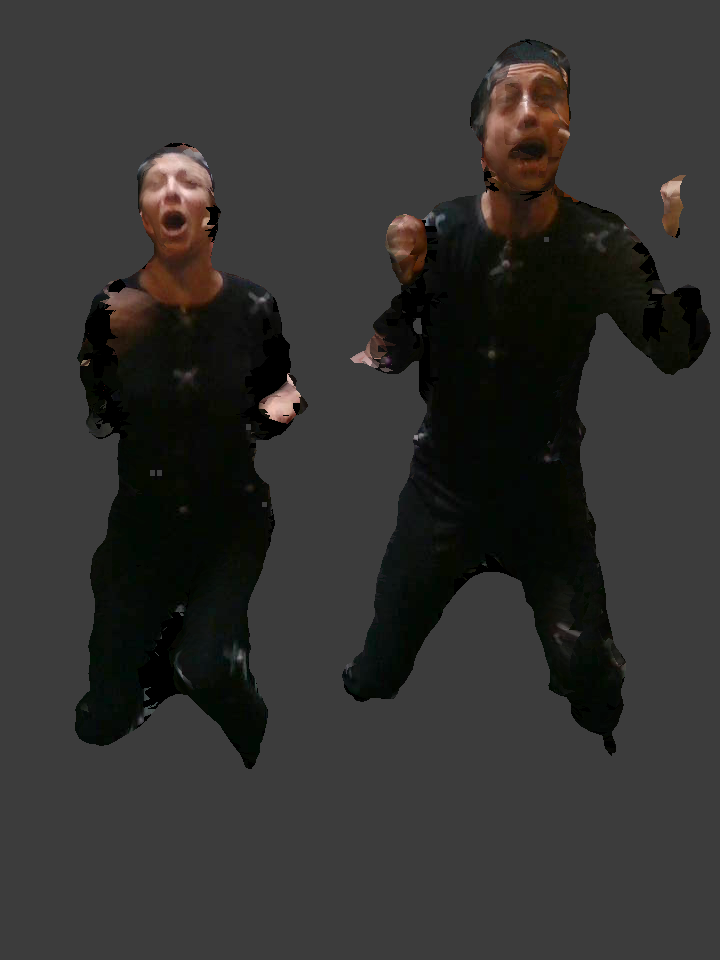}
\caption{$r=6$}
\end{subfigure}%
\begin{subfigure}{0.25\textwidth}
  \centering
  \includegraphics[width=1.0\linewidth]{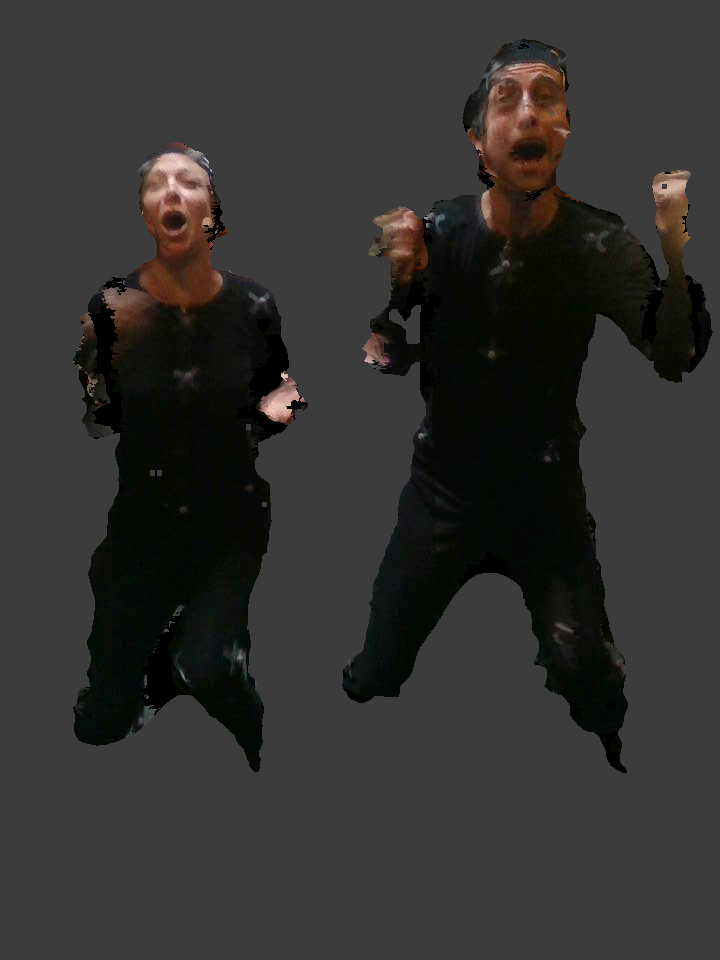}
\caption{$r=7$}
\end{subfigure}% 
\begin{subfigure}{0.25\textwidth}
  \centering
  \includegraphics[width=1.0\linewidth]{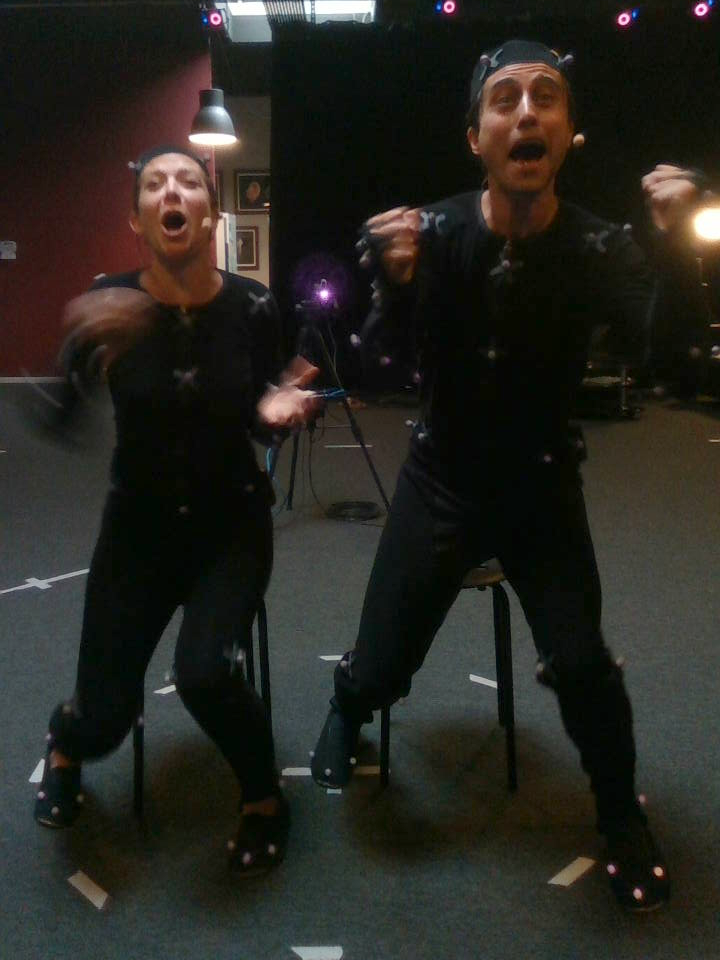}
\caption{\textit{RGB}}
\end{subfigure}
%%%%%%%%%%%%%%%%%%%
\caption{Textured mesh-based volumetric samples from H4D1 and H4D2 rendered in the 3 different voxel-grid resolutions along with the corresponding RGB images from the same viewpoint.}
\label{fig::drak_images}
\end{figure*}

\noindent \textbf{Results.} As can be seen in Tables \ref{fig::drak_single} and \ref{fig::drak_multi}, the experiments conducted, validate the claim that increments of a textured mesh voxel-grid resolution lead to increases in its objective visual quality.
Both for single- and multi-person evaluation sets, \textit{PSNR} increases in par with mesh resolution.
From $r=5$ to $r=6$ the increase is more pronounced, while from $r=6$ to $r=7$, it seems to diminish, indicating that a further increase in 3D mesh voxel-grid resolution may be futile, at least as regards the texture fidelity in terms of \textit{PSNR}.
The \textit{SSIM} case generally follows the same trend, with the exception of the \textit{S3} and \textit{S4} subjects from the single-person subset, where post $r=6$ increase in resolution does not seem to further improve the \textit{SSIM} of the textures.
In these cases, the $r=6$ and $r=7$ \textit{SSIM} values are approximately equal, exhibiting a difference of less than $10^{-4}$.

In Fig. \ref{fig::drak_images}, volumetric samples from the single- and multi-person subsets are illustrated, rendered in the 3 different voxel-grid resolutions along with the corresponding RGB images from the same viewpoint.
The increase of texture quality we want to highlight in these views is most apparent in the eyes area of the multi-person renderings.
As can be seen, for $r=5$ the right eye of the male subject is blurry and barely visible.
As the voxel-grid resolution increases, the eye gets crisper and better defined.
Such behaviour can be noticed in other areas of the volumetric data as well.

In a nutshell, experimental results indicate that the increase of 3D mesh voxel-grid resolution indeed leads to objective quality increase, though with diminishing returns.
This latter observation, together with the near real-time capabilities of the mesh-based volumetric reconstruction pipeline for $r=6$ and the decreased bandwidth needs it requires when compared with the $r=7$ case, makes $r=6$ voxel-grid resolution the most sensible choice for a volumetric live-streaming setup.

\section{Discussion}\label{sec::discussion}
We created HUMAN4D to provide the research community with a public resource that fills identified gaps in publicly available human-centric 4D datasets, consisting of motion  capture and HW-SYNCed volumetric data.
In the flood of recent literature, a plethora of algorithms and deep models focus on 3D pose estimation, however, only a few methods approach the task with the use of multi-view depth and volumetric data.
That is probably due to the complexity and time-consuming setup of multi-view capturing settings as well as the lack of spatio-temporally aligned multi-view depth maps with ground-truth data.
To this end, we aim to enable research on that direction encouraging the computer vision community to develop and experiment with new 3D pose estimation approaches on HUMAN4D by providing HW-SYNCed depth and volumetric data along with ultra-accurate ground-truth 3D poses for supervision and evaluation.
With regards to volumetric data, volumetric video is an emerging immersive medium, being unique due to its fully three-dimensional nature and its capability to enable six degrees of freedom (6DoF) spectating when used in 4D environments.
HUMAN4D has been created on the principle to provide spatio-temporally aligned mRGBD data captured to produce point- and mesh-based volumetric videos, reconstructed and compressed respecting online encoding and steady bit-rates.
On top of that, in most public datasets, the temporal misalignment between the multiple color and depth streams adds extra noise to the already noisy depth and color data, reducing the quality of the volumetric video.
In HUMAN4D, this noise is absent due to the high synchronization precision (HW-SYNC).

\section{Conclusion}\label{sec::conclusion}
In this paper we introduced HUMAN4D, a new multimodal human-centric 4D dataset containing a large corpus with more than 50K samples from daily, physical and social activities of annotated spatio-temporally aligned multi-view RGBD, volumetric and motion capture data along with audio recordings. 
To the best of our knowledge, HUMAN4D is the first dataset that provides HW-SYNCed mRGBD frames with the use of recent consumer-grade depth sensing devices.
We also provide evaluation benchmarks based on discriminative pose estimation and volumetric data compression methods.
We make all the data\footnote{\url{http://dx.doi.org/10.21227/xjzb-4y45}} and code\footnote{\url{https://github.com/tofis/human4d_dataset}} available online, including the respective synchronization, calibration and camera parameters, along with data loaders and other processing, visualization and evaluation tools, for academic use and further research. 
In that scope, the authors commit to continuously maintain the dataset for the community by adding new tools, baselines and captures.
Despite the continuous maintenance of the dataset, benchmarking subsets will remain constant to allow the assessment and comparison between new state-of-the-art methods on the same datasets.
We believe that HUMAN4D and its associated tools will stimulate further research in computer vision and data driven approaches, enabling research on human pose estimation, real-time volumetric video reconstruction and compression, with the use of consumer-grade RGBD cameras sensors.

\section{Acknowledgements}\label{sec::acknowledgements}

We gratefully appreciate the work conducted by the team of the Artanim Foundation Motion Capture Studio, providing high quality motion capture and 3D scanning services. 
We also want to give special thanks to Sylvain Chagué and Valérie Juillard, members of Artanim team, for scanning, post-processing and rigging of the 3D character and for post-processing and retargeting of the animations, respectively. 
Finally, we also acknowledge financial support by the H2020 EC project VRTogether under
contract 762111.

\vspace{6pt} 

\bibliography{bib}{}
\bibliographystyle{IEEEtran}

\begin{IEEEbiography}[{\includegraphics[width=1.1in,height=1.5in,clip,keepaspectratio]{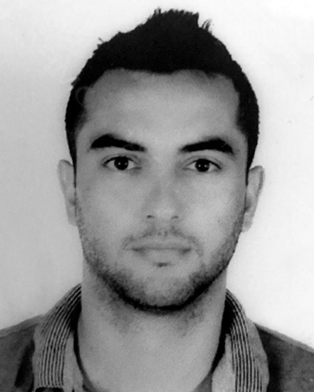}}]{\textbf{Anargyros Chatzitofis}} SIEEE, received his diploma in Electrical \& Computer Engineering (ECE) from ECE of National Technical University of Athens (NTUA). 
His main research expertise lies on human-centric 3D vision \& machine learning.
His PhD research at NTUA ECE is focused on depth-based motion capture \& deep learning.
He has (co)-authored more than 20 scientific publications in international Computer Vision conferences \& journals.
\end{IEEEbiography}
\begin{IEEEbiography}[{\includegraphics[width=1.1in,height=1.5in,clip,keepaspectratio]{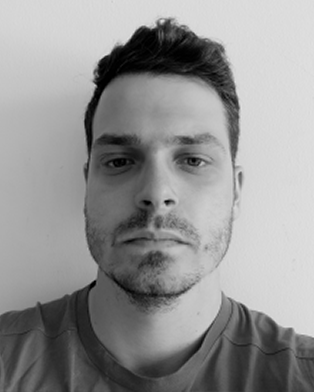}}]{\textbf{Leonidas Saroglou}} graduated from the department of Electrical \& Computer Engineering of Aristotle University of Thessaloniki (A.U.Th.). Since then, he has been working as a research assistant at the Information Technologies Institute (I.T.I.) of Centre for Research \& Technology Hellas (CERTH). His research includes image processing, pattern recognition, real-time 3D reconstruction, 3D computer vision \& deep learning.
\end{IEEEbiography}
\begin{IEEEbiography}[{\includegraphics[width=1.1in,height=1.5in,clip,keepaspectratio]{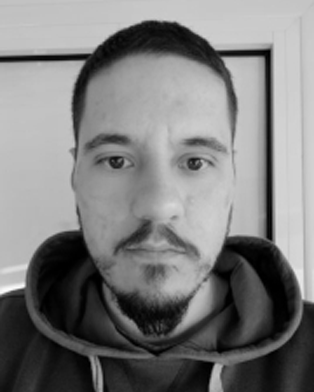}}]{\textbf{Prodromos Boutis}} graduated from the department of Electrical \& Computer Engineering of Aristotle University of Thessaloniki (A.U.Th.) in July 2018. 
Since February 2019, he has been working as a research assistant at the Information Technologies Institute (I.T.I.) of Centre for Technological Research \& Technology Hellas (CERTH). 
His research interests are 3D computer vision, digital image processing, 3D model rendering, pattern recognition \& machine learning.
\end{IEEEbiography}
\begin{IEEEbiography}[{\includegraphics[width=1.1in,height=1.5in,clip,keepaspectratio]{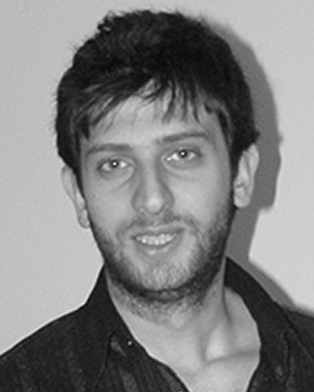}}]{\textbf{Petros Drakoulis}} received his BSc in IT Engineering from Alexander TEI \& his MSc (hons) in Digital Media \& Computational Intelligence from Aristotle University of Thessaloniki. In 2018, he joined the Visual Computing Laboratory of ITI-CERTH where he works as a Research Assistant \& Software Developer ever since. His main areas of interest include Software Engineering, Visual Computing, Machine Learning \& Graphics.
\end{IEEEbiography}
\begin{IEEEbiography}[{\includegraphics[width=1.1in,height=1.5in,clip,keepaspectratio]{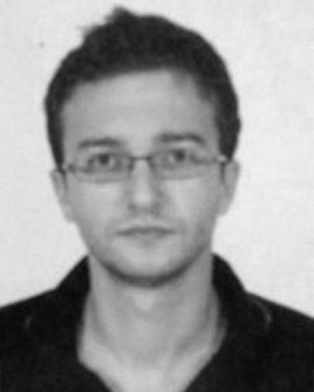}}]{\textbf{Nikolaos Zioulis}} is an Electrical \& Computer Engineer (Aristotle University of Thessaloniki, 2012) working in the Information Technologies Institute (ITI) of the Centre for Research \& Technology Hellas (CERTH) since October 2013. 
His research interests lie in the intersection of computer vision \& graphics technologies and, more specifically, in volumetric 3D capturing \& rendering, 3D scene understanding \& tele-immersive applications.
%He has a solid software engineering background \& is therefore keen on anything related to performance oriented real-time computer vision \& graphics.
\end{IEEEbiography}
\begin{IEEEbiography}[{\includegraphics[width=1.1in,height=1.5in,clip,keepaspectratio]{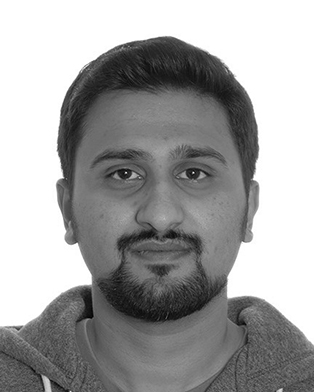}}]{\textbf{Shishir Subramanyam}} received his BTech in Computer Science from BITS Pilani Dubai \& his MSc in Computer Science from Delft University of Technology. He is currently a PhD candidate at Centrum Wiskunde \& Informatica with the Distributed \& Interactive Systems Group. His research interests are on Multimedia Systems specifically on the transport \& delivery of volumetric media.
\end{IEEEbiography}
\begin{IEEEbiography}[{\includegraphics[width=1.1in,height=1.5in,clip,keepaspectratio]{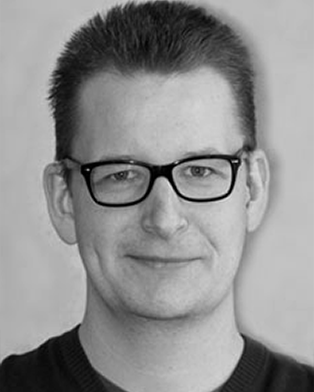}}]{\textbf{Bart Kevelham}} received his MSc in Computer Science in 2006 from the University of Twente, the Netherlands, specializing in computer graphics. He currently is Lead R\&D Engineer at Artanim in Geneva Switzerland, where his work focuses on the research \& development of solutions enabling interactive full-body \& free-roam VR experiences. His research interests include real-time Computer Graphics, Physical Simulation \& Computer Vision. 
\end{IEEEbiography}
\begin{IEEEbiography}[{\includegraphics[width=1.1in,height=1.5in,clip,keepaspectratio]{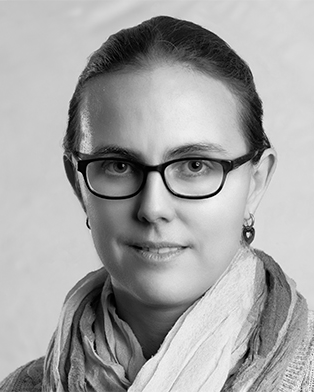}}]{\textbf{Caecilia Charbonnier}} obtained a PhD degree in Computer Science in 2010 at MIRALab - University of Geneva, Switzerland. She is the Co-Founder \& Research Director of Artanim, a center specialized in motion capture technologies, \& Co-Founder \& CIO of Dreamscape Immersive, a VR entertainment company. Her work focus on the interdisciplinary use of motion capture from 3D animation, live performances to movement science, orthopedics \& sports medicine.
\end{IEEEbiography}
\begin{IEEEbiography}[{\includegraphics[width=1.1in,height=1.5in,clip,keepaspectratio]{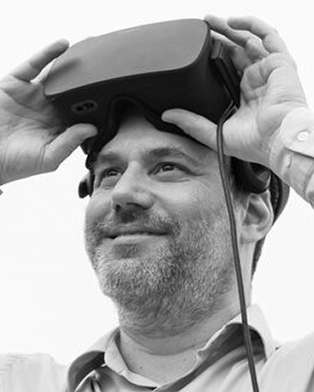}}]{\textbf{Pablo Cesar}} leads the Distributed \& Interactive Systems Group, Centrum Wiskunde \& Informatica (CWI) \& is Associate Professor with TU Delft, The Netherlands. 
His research combines HCI \& multimedia systems, \& focuses on modelling \& controlling complex collections of media objects distributed in time \& space. 
He is a member of the Editorial Board of IEEE Multimedia, ACM Transactions on Multimedia, \& IEEE Transactions of Multimedia, among others.
\end{IEEEbiography}
\begin{IEEEbiography}[{\includegraphics[width=1.1in,height=1.5in,clip,keepaspectratio]{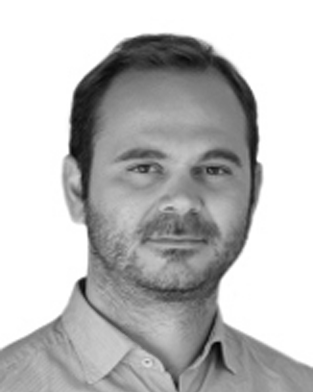}}]{\textbf{Dimitrios Zarpalas}} is a senior researcher (Grade C) at Information Technologies Institute (ITI) of Centre for Research \& Technology Hellas (CERTH). He holds the diploma of Electrical \& Computer Engineer from Aristotle University of Thessaloniki, A.U.Th, an MSc in computer vision from The Pennsylvania State University, \& a PhD in medical informatics (Health Science School, department of Medicine, A.U.Th). He joined ITI in 2008, as an Associate Researcher.
\end{IEEEbiography}
\begin{IEEEbiography}[{\includegraphics[width=1.1in,height=1.5in,clip,keepaspectratio]{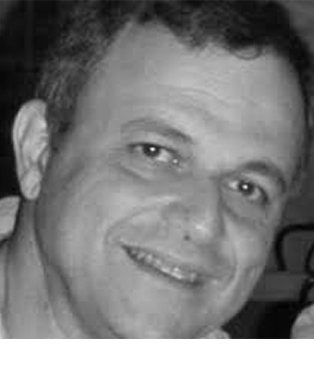}}]{\textbf{Stefanos Kollias}}, FIEEE, FHEA, has been Professor in ECE School of the National Technical University of Athens, since 1997.
He has been Professor of Machine Learning in the Computer Science School of the University of Lincoln, UK, since 2016. His research covers machine \& deep learning, multimedia analysis, search, retrieval \& recognition, vision, medical informatics, cultural heritage, HCI \& affective computing.
He has published 110 journal papers \& 310 conference papers.
He has supervised 43 Ph.D. students. 
\end{IEEEbiography}
\begin{IEEEbiography}[{\includegraphics[width=1.1in,height=1.5in,clip,keepaspectratio]{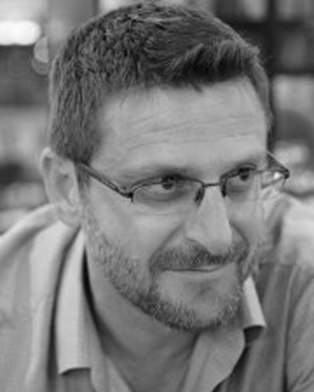}}]{\textbf{Petros Daras}} received the Diploma in electrical \& computer engineering \& the M.Sc. \& Ph.D. degrees in electrical \& computer engineering from the Aristotle University of Thessaloniki, Greece, in 1999, 2002, \& 2005, respectively. 
He is currently a Research Director \& the Chair of the Visual Computing Lab coordinating the research effort of more than 80 scientists \& engineers. 
His main research interests include visual content processing, multimedia indexing \& machine learning. 
His involvement with those research areas has led to the coauthoring of more than 300 articles in refereed journals \& international conferences. 
\end{IEEEbiography}
\EOD

\end{document}